\documentclass{article}

\usepackage[final]{neurips_2024}

\usepackage[utf8]{inputenc} 
\usepackage[T1]{fontenc}    
\usepackage{hyperref}       
\usepackage{url}            
\usepackage{booktabs}       
\usepackage{nicefrac}       
\usepackage{microtype}      
\usepackage{xcolor}         

\usepackage{amsmath,amssymb,amsfonts}
\usepackage{bm}
\usepackage{algorithmic}
\usepackage{graphicx}
\usepackage{textcomp}
\usepackage{xcolor}

\usepackage{floatrow}
\newfloatcommand{capbtabbox}{table}[][\FBwidth]
\usepackage{wrapfig}

\usepackage{dsfont}
\usepackage{algorithm}
\usepackage{subfigure}
\usepackage{booktabs}
\usepackage{balance}
\usepackage{comment}
\usepackage{url}
\usepackage{color}
\usepackage{array}
\usepackage{multirow}
\usepackage{hyperref}
\newcommand{\red}{\color{red}}
\newcommand{\blue}{\color{blue}}

\title{A Consistency-Aware Spot-Guided Transformer for Versatile and Hierarchical Point Cloud Registration}

\author{
  Renlang Huang\hspace{0.7cm} Yufan Tang\hspace{0.7cm} Jiming Chen\hspace{0.7cm} Liang Li \\
  College of Control Science and Engineering\\
  Zhejiang University, Hangzhou 310027, China \\
  \texttt{\{renlanghuang,tyfan,cjm,liang.li\}@zju.edu.cn} \\
}

\begin{document}

\maketitle

\begin{abstract}
Deep learning-based feature matching has shown great superiority for point cloud registration in the absence of pose priors. Although coarse-to-fine matching approaches are prevalent, the coarse matching of existing methods is typically sparse and loose without consideration of geometric consistency, which makes the subsequent fine matching rely on ineffective optimal transport and hypothesis-and-selection methods for consistency.
Therefore, these methods are neither efficient nor scalable for real-time applications such as odometry in robotics. To address these issues, we design a consistency-aware spot-guided Transformer (CAST), which incorporates a spot-guided cross-attention module to avoid interfering with irrelevant areas, and a consistency-aware self-attention module to enhance matching capabilities with geometrically consistent correspondences. Furthermore, a lightweight fine matching module for both sparse keypoints and dense features can estimate the transformation accurately. Extensive experiments on both outdoor LiDAR point cloud datasets and indoor RGBD point cloud datasets demonstrate that our method achieves \textit{state-of-the-art} accuracy, efficiency, and robustness.
Our code is available at \url{https://github.com/RenlangHuang/CAST}.
\end{abstract}

\section{Introduction}

Point cloud registration is a fundamental yet crucial task for a variety of 3D vision and robotic applications, such as simultaneous localization and mapping (SLAM)~\cite{huang2024kdd}, object pose estimation~\cite{hua2021object} and structure from motion (SfM)~\cite{wang2021sfm}.
Aiming at aligning two partially overlapped point clouds, the typical approach involves a two-stage pipeline: data association which establishes reliable point correspondences, and pose estimation. However, establishing these correspondences has been challenging due to the noisy, irregular, non-uniform, and textureless nature of 3D point clouds.

Feature matching has long been the mainstream of data association without pose priors.
Extensive research has made advances in distinctive local feature representations, ranging from hand-crafted descriptors~\cite{rusu2008pfh, rusu2009fpfh, salti2014shot} to recent learning-based descriptors~\cite{gojcic2019perfect, choy2019fcgf, bai2020d3feat}.
Although the emerging learning-based descriptors significantly improve the reliability of correspondences, the inlier ratio still falls short of what is required for robust and efficient pose estimation.
Recently, coarse-to-fine matching is a thriving framework for 2D-2D~\cite{yu2023astr}, 3D-3D~\cite{yu2021cofinet,qin2023geotransformer,yang2022one}, and even 2D-3D~\cite{li20232d3d} data association. 
It has been a consensus that Transformers stacked by alternate self-attention and cross-attention modules are effective for coarse matching, which are inspired by human visual processes.
Typically, humans may first scan through the point clouds to identify and match salient landmarks across different point clouds reliably.
For less salient points, the geometric relationships between them and those salient landmarks would be utilized to revisit their potential correspondences. 
The correspondences will eventually be established for the entire point cloud after several iterations of this process.

Unfortunately, existing coarse matching approaches tend to be sparse and loose without consideration of geometric consistency.
An important reason for looseness is that global cross-attention inevitably attends to similar yet irrelevant areas, resulting in misleading feature aggregation and consequent inconsistent correspondences that undermine both robustness and accuracy.
As a result, the hypothesis-and-selection pipeline such as RANSAC~\cite{fischler1981ransac} is commonly used for outlier rejection, which is typically inaccurate and inefficient, especially for numerous samples with low inlier ratio.
Furthermore, the sparsity necessitates the use of complicated fine matching such as optimal transport-based algorithms to establish reliable dense correspondences. Due to iterative dense matrix operations for patch-to-patch correspondences established by coarse matching, these fine matching methods are neither efficient nor scalable for real-time large-scale applications such as odometry.

To this end, we attempt to design an efficient and scalable coarse-to-fine matching network based on consistency-aware semi-dense coarse correspondences.
Inspired by ASTR~\cite{yu2023astr} for 2D feature matching, we leverage local consistency to direct the cross-attention of each point exclusively to corresponding patches of its confident neighbors, which is referred to as spot-guided cross-attention.
Unlike~\cite{yu2023astr}, we propose a novel consistency-aware matching confidence criterion to sample reliable neighbors based on both feature similarity and geometric compatibility.
Additionally, we design a consistency-aware self-attention module to enhance the distinctiveness of coarse feature representations via aggregation with salient nodes from the compatibility graph.
Notably, both spot-guided cross-attention and consistency-aware self-attention are efficient sparse attention mechanisms.

For scalability to real-time applications such as odometry with pose priors, we propose a lightweight fine matching module allowing independent deployment without coarse matching.
The scalability is credited to flexible point-to-patch local matching instead of optimal transport heavily relying on patch-to-patch correspondences.
In addition, our fine matching adopts a sparse-to-dense registration pipeline, benefiting from the efficiency of sparse keypoint matching and the accuracy of dense registration.
Furthermore, an efficient compatibility graph embedding module is leveraged for outlier rejection as a substitute for inefficient hypothesis-and-selection pipelines.

In summary, our main contributions are as follows:
\begin{itemize}
\item A consistency-aware spot-guided Transformer (CAST) with multi-scale feature fusion for much tighter coarse matching with a focus on geometric consistency.
\item A spot-guided cross-attention module with a consistency-aware matching confidence criterion that can maintain local consistency without interfering with irrelevant areas.
\item A consistency-aware self-attention module based on sparse sampling from the compatibility graph to enhance global consistency during feature aggregation.
\item A lightweight and scalable sparse-to-dense matching module involving both sparse keypoints and dense features to achieve lower registration errors without optimal transport and hypothesis-and-selection pipelines.
\end{itemize}

\section{Related Work}

\paragraph{3D Feature Descriptors.}
Feature matching plays a crucial role in point cloud registration, enabling the establishment of reliable correspondences without pose priors.
Early methods use hand-crafted descriptors based on signatures~\cite{salti2014shot} or histograms~\cite{rusu2008pfh,rusu2009fpfh} to represent local geometric features.
Recently, learning-based 3D descriptors have showcased greater performance than hand-crafted ones, which are usually trained in a self-supervised manner by maximizing the similarity between descriptors of true correspondences and minimizing the similarity otherwise.
3DMatch~\cite{zeng20173dmatch} and PerfectMatch~\cite{gojcic2019perfect} leverage 3D CNNs to learn local patch-wise descriptors from 3D patches converted into voxels of truncated distance function (TDF) values and smoothed density value (SDV) representations, respectively.
PPFNet~\cite{deng2018ppfnet} extracts global context-aware patch-wise descriptors based on PointNet~\cite{qi2017pointnet}.
FCGF~\cite{choy2019fcgf} employs a sparse 3D convolutional encoder-decoder network for dense descriptor learning.
SpinNet~\cite{ao2021spinnet} proposes a 3D cylindrical convolution network to extract rotation-invariant patch-wise descriptors. Predator~\cite{huang2021predator} utilizes graph convolution and cross-attention to enhance the descriptors and predict the overlapping regions for robust performance in low overlap scenarios.

\paragraph{3D Keypoint Detectors.}
Detection-based methods have been widely studied in image matching but less developed for 3D point clouds. 
Existing 3D keypoint detectors are mainly hand-crafted, which extract salient points based on unique geometric features such as specific curvatures~\cite{chen2007det} or principal directions~\cite{zhong2009iss}. However, they suffer from noisy, sparse, and non-uniform real-world point clouds with large-scale transformations.
Recent advances include learning-based detectors such as USIP~\cite{li2019usip} that predicts repeatable keypoints by minimizing a probabilistic chamfer loss, and HRegNet~\cite{lu2023hregnet} that further utilizes weighted farthest point sampling to select sparse keypoints from the predicted ones for hierarchical registration.
3DFeat-Net~\cite{yew20183dfeat} extracts patch-wise descriptors with saliency scores for keypoint selection in a weakly supervised manner by minimizing a weighted feature alignment triplet loss.
D3Feat~\cite{bai2020d3feat} adopts a fully convolutional network to predict point-wise descriptors with hand-crafted saliency scores by minimizing a self-supervised detection loss.

\paragraph{3D Correspondence Learning.}
DCP~\cite{wang2019dcp} predicts soft correspondences from learned features and estimates the pose by a differential SVD layer.
IDAM~\cite{li2020idam} designs iterative distance-aware similarity matrix convolution for iterative pairwise matching and pose estimation.
Recently, coarse-to-fine correspondence learning has been regarded as a promising approach.
The pioneering work CoFiNet~\cite{yu2021cofinet} exploits a group of self-attention and cross-attention for coarse feature matching and the optimal transport for fine matching. 
GeoTransformer~\cite{qin2023geotransformer} proposes a geometric structure embedding for self-attention, and the local-to-global registration (LGR) for consistent pose estimation.
RoITr~\cite{yu2023roitr} improves the coarse-to-fine framework with a rotation-invariant point cloud Transformer based on point pair features, while PEAL~\cite{yu2023peal} and DiffusionPCR~\cite{chen2023diffusionpcr} use overlap priors and diffusion models for iterative feature matching, respectively.
For outlier rejection, RANSAC~\cite{fischler1981ransac} remains popular despite its inefficiency.
DGR~\cite{choy2020dgr} predicts correspondence-wise confidence scores via a 6D convolutional network, while PointDSC~\cite{bai2021pointdsc} designs a consistency-guided non-local feature embedding to sample consistent correspondences for neural spectral matching and pose estimation.

\section{Method}
\label{sec:method}

In this section, we present the proposed consistency-aware spot-guided Transformer (CAST) with a lightweight sparse-to-dense fine matching module for accurate and efficient point cloud registration.

\subsection{Overview}
Given two partially overlapped point clouds $\mathbf{X}=\{\mathbf{x}_i\in\mathbb{R}^3,i=1,\cdots, M\}$ and $\mathbf{Y}=\{\mathbf{y}_j\in\mathbb{R}^3,j=1,\cdots, N\}$, the point cloud registration problem can be formulated as solving the optimal rigid transformation between $\mathbf{X},\mathbf{Y}$ by minimizing the weighted sum of point-to-point errors of a predicted correspondence set $\mathcal{C}$ with a confidence weight $w_k$ for each correspondence $(\mathbf{x}_k,\mathbf{y}_k)$:
\begin{equation}
\label{eq:registration}
\min_{\mathbf{R},\mathbf{t}}\sum_{(\mathbf{x}_k,\mathbf{y}_k)\in\mathcal{C}} w_k\Vert \mathbf{R}\mathbf{x}_k + \mathbf{t} - \mathbf{y}_k \Vert_2^2,
\end{equation}
where $\mathbf{R}\in SO(3)$ and $\mathbf{t}\in\mathbb{R}^3$ are the rotation and the translation between $\mathbf{X}$ and $\mathbf{Y}$, respectively.

\begin{figure}[t]
\centering
\includegraphics[width=\linewidth]{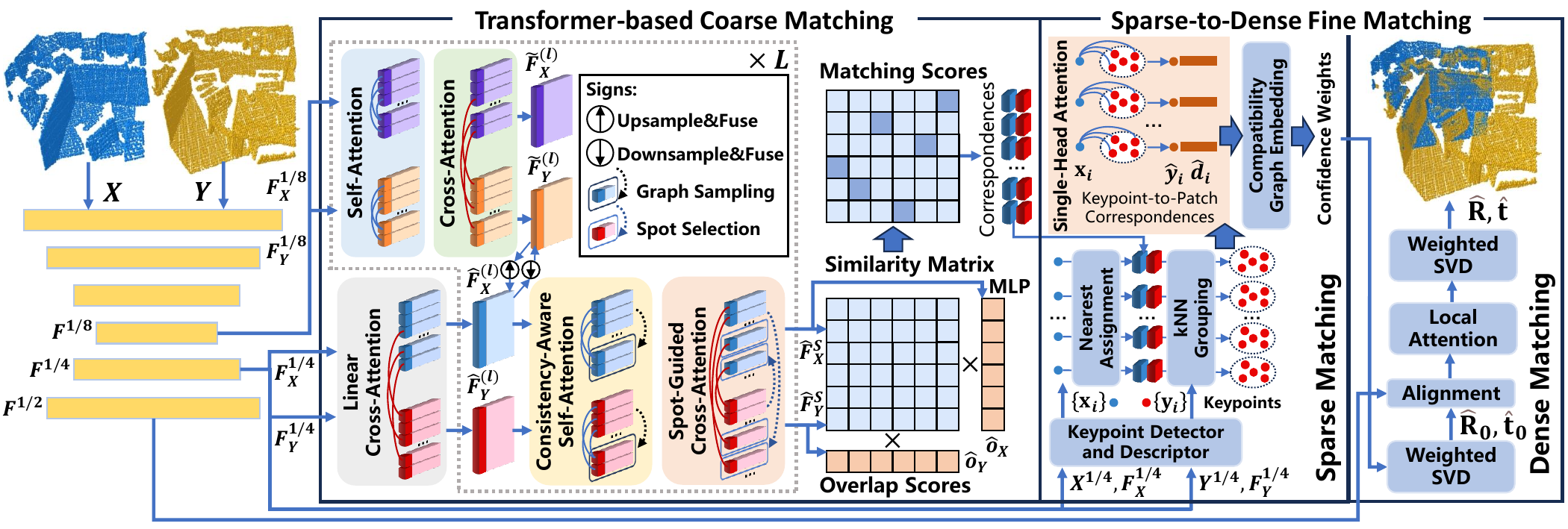}
\caption{Overview of CAST.
The feature pyramid network down-samples the point clouds and learns features in multiple resolutions. The coarse matching module extracts consistency-aware semi-dense correspondences via a group of alternate consistency-aware self-attention modules and spot-guided cross-attention modules with multi-scale feature fusion. Finally, the fine matching module predicts correspondences for both sparse keypoints and dense features and estimates the transformation.
}
\label{fig:overview}
\end{figure}

As depicted in Figure~\ref{fig:overview}, CAST follows a coarse-to-fine feature matching and registration architecture, including a feature pyramid network, a consistency-aware spot-guided attention-based coarse matching module, and a sparse-to-dense fine matching module.
We first utilize a KPConv-based fully convolutional network~\cite{thomas2019kpconv} to extract multi-scale features.
We denote feature maps of the decoder with the size of $1/k$ as $\mathbf{F}^{1/k} = \{\mathbf{F}^{1/k}_X, \mathbf{F}^{1/k}_Y\}$, which correspond to nodes $\mathbf{X}^{1/k}$ and $\mathbf{Y}^{1/k}$ down-sampled from $\mathbf{X}, \mathbf{Y}$, respectively.
For coarse matching, we first adopt an efficient linear cross-attention~\cite{katharopoulos2020transformers} module to enhance $\mathbf{F}^{1/4}$.
Then both \textit{semi-dense features} $\mathbf{F}^{1/4}$ and \textit{coarse features} $\mathbf{F}^{1/8}$ are fed into a consistency-aware spot-guided attention-based coarse matching module to improve the feature distinctiveness.
The similarity matrix $\mathbf{S}\in \mathbb{R}^{M'\times N'}$ between these enhanced semi-dense features $\mathbf{\hat{F}}_{X}\in \mathbb{R}^{M'\times D},\mathbf{\hat{F}}_{Y}\in \mathbb{R}^{N'\times D}$ is computed based on inner product: $\mathbf{S}=\mathbf{\hat{F}}_{X} \mathbf{\hat{F}}_{Y}^{\mathsf{T}}$.
Furthermore, we fed $\mathbf{\hat{F}}_{X}$ and $\mathbf{\hat{F}}_{Y}$ into a point-wise MLP to predict the overlap scores, which encode the likelihood of a node having a correspondence.
We perform dual-softmax on $\mathbf{S}$ to obtain the final matching scores:
\begin{equation}
\label{eq:coarse}
\mathbf{P}_{ij} = \hat{o}_i^X\hat{o}_j^Y \underset{k\in \{1,\cdots,M'\}}{\text{softmax}}(\mathbf{S}_{kj})_i \underset{k\in \{1,\cdots,N'\}}{\text{softmax}}(\mathbf{S}_{ik})_j,
\end{equation}
where $\hat{o}_i^X$ and $\hat{o}_j^Y$ are predicted overlap scores of the \textit{i}-th node of $\mathbf{X}^{1/4}$ and the \textit{j}-th node of $\mathbf{Y}^{1/4}$, respectively.
We use the mutual nearest neighbor scheme to select confident coarse correspondences.
For efficient fine matching, we extract a keypoint from the neighborhood of each semi-dense node in $\mathbf{X}^{1/4}$, and predict its virtual correspondence in $\mathbf{Y}^{1/4}$ based on the lightweight single-head attention.
Then we utilize compatibility graph embedding to predict the confidence of these keypoint correspondences as weights in Eq.~\ref{eq:registration} for initial pose estimation.
Finally, a lightweight local attention module for dense points $\mathbf{X}^{1/2}$ and $\mathbf{Y}^{1/2}$ predicts dense correspondences to refine the pose.

\subsection{Consistency-Aware Spot-Guided Attention}
To tackle the sparsity and looseness of coarse matching, we focus on feature aggregation among semi-dense features $\mathbf{F}^{1/4}$ leveraging both local and global geometric consistency. To be specific, the self-attention only attends to salient nodes sampled from a global compatibility graph, while the cross-attention only attends to nodes sampled based on local consistency, which are referred to as \textit{consistency-aware self-attention} and \textit{spot-guided cross-attention}, respectively.

\paragraph{Preliminaries.}
Transformers stacked by alternate self-attention and cross-attention have showcased advanced performance in coarse feature matching. When $D$-dimensional features $\mathbf{F}_A$ attends to $\mathbf{F}_B$, the output of vanilla attention  is formulated as:
\begin{equation}
\label{eq:cross-attention}
\hat{\mathbf{F}}_A = \text{softmax}\left( \frac{1}{\sqrt{D}}\mathbf{F}_A\mathbf{W}_Q(\mathbf{F}_B\mathbf{W}_K)^\mathsf{T}\right)\mathbf{F}_B\mathbf{W}_V,
\end{equation}
where $\mathbf{W}_Q,\mathbf{W}_K,\mathbf{W}_V$ are learnable linear transformations to generate queries, keys, and values. When $\mathbf{F}_A,\mathbf{F}_B$ related to coordinates $\mathbf{P}_A,\mathbf{P}_B$ are from the same point cloud, it becomes self-attention that requires positional encoding to embed spatial information. To encode the 3D relative positions, we equip the rotary positional embedding~\cite{su2024roformer} $\tilde{\mathbf{R}}(\cdot)$ with learnable weights $\mathbf{b}_1,\cdots,\mathbf{b}_{D/2}\in\mathbb{R}^{1\times 3}$:
\begin{equation}
\label{eq:rotary}
\tilde{\mathbf{R}}(\mathbf{p}) = \begin{bmatrix}
    \mathbf{R}(\mathbf{b}_1\mathbf{p}) & & \mathbf{0}\\
    & \ddots &\\
    \mathbf{0} & & \mathbf{R}(\mathbf{b}_{D/2}\mathbf{p})
\end{bmatrix},
\mathbf{R}(\theta) = \begin{bmatrix}
  \cos{\theta} & -\sin{\theta} \\
  \sin{\theta} & \cos{\theta}
\end{bmatrix},\forall \mathbf{p}\in\mathbb{R}^3.
\end{equation}
When applying $\tilde{\mathbf{R}}(\cdot)$ to vanilla self-attention, the output is formulated as:
\begin{equation}
\label{eq:self-attention}
\hat{\mathbf{F}}_A = \text{softmax}\left( \frac{1}{\sqrt{D}}\mathbf{F}_A\mathbf{W}_Q\tilde{\mathbf{R}}(\mathbf{P}_A)(\mathbf{F}_B\mathbf{W}_K\tilde{\mathbf{R}}(\mathbf{P}_B))^\mathsf{T}\right)\mathbf{F}_B\mathbf{W}_V.
\end{equation}

\paragraph{Architecture.}
As both spot-guided cross-attention and consistency-aware self-attention are sparse attention lacking of abundant global context, we propose to enhance the semi-dense features via multi-scale feature fusion with coarse features.
Hence, the architecture of our coarse matching module is designed as a sequence of blocks for attention-based multi-scale feature aggregation.
For each block with both semi-dense features $\mathbf{F}^{1/4}$ and coarse features $\mathbf{F}^{1/8}$ as inputs, we first feed $\mathbf{F}^{1/8}$ into a self-attention module (Eq.~\ref{eq:self-attention}) and a cross-attention module (Eq.~\ref{eq:cross-attention}).
Then $\mathbf{F}^{1/4}$ and $\mathbf{F}^{1/8}$ are fused into each other based on nearest up-sampling and distance-based interpolated down-sampling~\cite{qi2017pointnet}:
\begin{equation}
\begin{aligned}
\mathbf{\hat{F}}^{1/4} &= \mathbf{F}^{1/4} + \text{MLP}(\text{Nearest Up-sampling}(\mathbf{F}^{1/8})), \\
\mathbf{\hat{F}}^{1/8} &= \mathbf{F}^{1/8} + \text{MLP}(\text{Interpolated Down-sampling}(\mathbf{F}^{1/4})).
\end{aligned}
\end{equation}
Finally, $\mathbf{\hat{F}}^{1/4}$ is fed into a consistency-aware self-attention module and a spot-guided cross-attention module at the end of each block. 
Before these sparse attention modules, we need to match the semi-dense features and evaluate the geometric consistency as a clue to select sparse yet instructive tokens. Given semi-dense features $\mathbf{\hat{F}}_{X}^{(l)},\mathbf{\hat{F}}_{Y}^{(l)}$ in the $l$-th block, the matching score is formulated as:
\begin{equation}
\label{eq:match}
\mathbf{P}_{ij}^{(l)} = \underset{k\in \{1,\cdots,M'\}}{\text{softmax}}(\mathbf{S}_{kj}^{(l)})_i \underset{k\in \{1,\cdots,N'\}}{\text{softmax}}(\mathbf{S}_{ik}^{(l)})_j,\;
\mathbf{S}^{(l)}=\mathbf{\hat{F}}_{X}^{(l)} (\mathbf{\hat{F}}_{Y}^{(l)})^{\mathsf{T}}.
\end{equation}
Then the correspondence of each node can be obtained as the node from another point cloud with the highest matching score, forming a correspondence set $\mathcal{C}^{(l)}=\{(\mathbf{x}_i^S,\mathbf{y}_i^S): \mathbf{x}_i^S\in \mathbf{X}^{1/4},\mathbf{y}_i^S\in\mathbf{Y}^{1/4}\}$.
An insight about the consistency among correspondences is that the distance between two points is invariant after transformation.
Hence, geometric compatibility is adopted as a simple yet effective measure of consistency~\cite{bai2021pointdsc}, which is based on the length difference between pairwise line segments. Given a pre-defined threshold $\sigma_c$, the pair-wise geometric compatibility of $\mathcal{C}^{(l)}$ is formulated as:
\begin{equation}
\label{eq:compatibility}
\beta_{ij}=\left[1 - d_{ij}^2 / \sigma_c^2 \right]^+,d_{ij}=\left| \Vert \mathbf{x}_i^S - \mathbf{x}_j^S\Vert_2 - \Vert \mathbf{y}_i^S - \mathbf{y}_j^S\Vert_2\right|.
\end{equation}
The compatibility matrix $\mathbf{B}_c=[\beta_{ij}]_{M'\times N'}$ is also considered as the adjacency matrix of a weighted undirected graph known as the compatibility graph, where each vertex is a pair of correspondence and the edge connectivity corresponds to the compatibility between two correspondences.
Intuitively, we adopt the generalized degree of a pair of correspondence in the graph as a measure of global consistency, which quantifies the connectivity of a vertex as the sum of edge weights connected to it.

\begin{figure}[t]
\centering
\includegraphics[width=\linewidth]{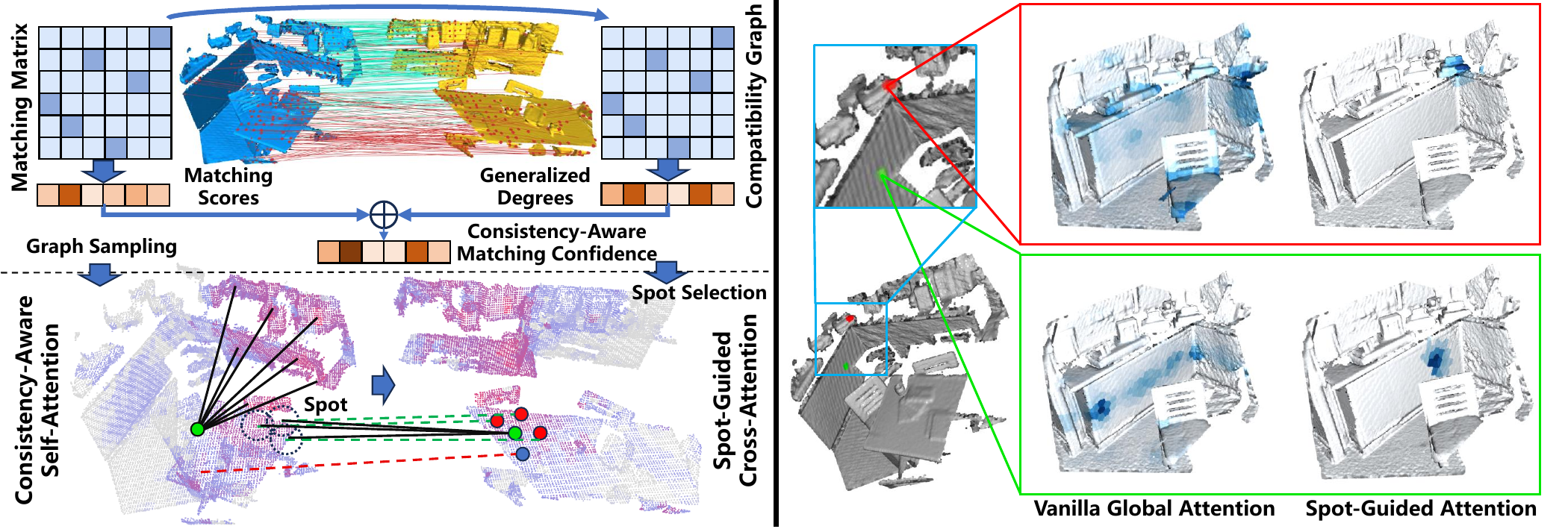}
\caption{Illustration of consistency-aware self-attention and spot-guided cross-attention (Left), as well as visualization of the global cross-attention and spot-guided cross-attention (Right). For the left part, the green nodes are query nodes, while the red ones with correct correspondences (green dot lines) are reliable neighbors, and the blue one with a false correspondence (red dot line) is an unreliable neighbor. The self-attention (black lines) only attends to salient nodes while the cross-attention (black lines) only attends to spots (nodes within black circles).}
\label{fig:attention}
\end{figure}

\paragraph{Consistency-Aware Self-Attention.}
Intuitively, the correspondences of less salient nodes can be effectively located based on the geometric relationships between them and the salient ones.
Hence, compared with global self-attention that attends to all nodes, attending to only salient nodes is more efficient and effective to encode the geometric context for matching.
We propose the consistency-aware self-attention that samples sparse salient nodes to be attended to based on both geometric consistency and feature similarity.
Given the correspondence set $\mathcal{C}^{(l)}$ with a compatibility graph, we perform two-stage sampling by ranking the generalized degrees and matching scores, respectively.
The first-stage graph sampling using generalized degrees can obtain sufficient consistent correspondences as proposals.
The second-stage sampling based on matching scores can further obtain sparse salient nodes from these proposals.
Finally, semi-dense features $\mathbf{\hat{F}}_{X}^{(l)},\mathbf{\hat{F}}_{Y}^{(l)}$ only attend to features of salient nodes from the same point cloud for feature aggregation according to Eq.~\ref{eq:self-attention}.

\paragraph{Spot-Guided Cross-Attention.}
As shown in Figure~\ref{fig:attention}, global cross-attention tends to aggregate features from many irrelevant regions with similar patterns, leading to false correspondences.
Inspired by local consistency that the correspondences of adjacent 3D points remain close to each other, we design the spot-guided cross-attention as depicted in Figure~\ref{fig:attention}.
For each node $\mathbf{x}_i^S$ such that $(\mathbf{x}_i^S,\mathbf{y}_i^S)\in \mathcal{C}^{(l)}$, we select a subset $\mathcal{N}_s(\mathbf{x}_i^S)$ from its neighborhood $\mathcal{N}(\mathbf{x}_i^S)$ as seeds, and construct a region of interest for it as 
$\mathcal{S}(\mathbf{x}_i^S)=\bigcup_{\mathbf{x}_k^S\in \mathcal{N}_s(\mathbf{x}_i^S)} \mathcal{N}(\mathbf{y}_k^S)$, namely its \textit{spot}.
$\mathcal{N}_s(\mathbf{x}_i^S)$ selects $\mathbf{x}_i^S$ and only its neighbors with reliable correspondences. We propose a consistency-aware matching confidence criterion to rank the neighbors, which is formulated as the product of the matching score and the normalized generalized degree in the compatibility graph.
This criterion incorporates feature similarity and geometric consistency to properly measure the reliability of correspondences for seed selection.
Finally, semi-dense features attend to their spots for feature aggregation according to Eq.~\ref{eq:cross-attention}.
Under the guarantee of local consistency, the spots are likely to cover the true correspondences, providing guidance for feature aggregation without interfering with irrelevant areas.

\subsection{Sparse-to-Dense Fine Matching}
Given a coarse correspondence set $\mathcal{\hat{C}}=\{(\mathbf{x}_j^S, \mathbf{y}_j^S): \mathbf{x}_j^S\in\mathbf{X}^{1/4}, \mathbf{y}_j^S\in\mathbf{Y}^{1/4}\}$ selected as mutual nearest neighbors from the final coarse matching scores (Eq.~\ref{eq:coarse}), we propose a lightweight sparse-to-dense fine matching module for hierarchical pose estimation without optimal transport, maintaining scalability and efficiency.
For sparse matching, we first search \textit{k}-nearest neighbors (kNN) of semi-dense nodes $\mathbf{X}^{1/4}$ among dense points $\mathbf{X}^{1/2}$ to group patches, then we use an attentive keypoint detector~\cite{lu2020rskdd} to predict a repeatable keypoint with a descriptor from each patch.
Each keypoint of point cloud $\mathbf{X}$ is assigned to its nearest node, and each node $\mathbf{y}_j^S\in\mathbf{Y}^{1/4}$ with a correspondence in $\mathcal{\hat{C}}$ groups a patch $\mathcal{P}(\mathbf{y}_j^S)$ of keypoints via kNN.
Then, a keypoint $\mathbf{x}_i$ assigned to $\mathbf{x}^S_j$ will correspond to the patch $\mathcal{P}(\mathbf{y}_j^S)$, forming a pair of keypoint-to-patch correspondence.
Finally, we utilize a shared single-head attention layer for each keypoint-to-patch correspondence to predict virtual correspondences for keypoints.
Denote the descriptor of $\mathbf{x}_i$ as $d^X_{i}$, the virtual correspondence $\hat{\mathbf{y}}_i$ with feature $\hat{d}_i^Y$ is predicted from keypoints $\mathbf{y}_{i_1},\cdots,\mathbf{y}_{i_k}$ with top-\textit{k} descriptor similarity in $\mathcal{P}(\mathbf{y}_j^S)$ as:
\begin{equation}
\label{eq:local}
\hat{\mathbf{y}}_i = \sum_{j=1}^k \text{softmax}(d^X_{i}\overline{\mathbf{W}}_Q(d^Y_{i_j}\overline{\mathbf{W}}_K)^\mathsf{T}) \mathbf{y}_{i_k}, \hat{d}_i^Y = \sum_{j=1}^k \text{softmax}(d^X_{i}\overline{\mathbf{W}}_Q(d^Y_{i_j}\overline{\mathbf{W}}_K)^\mathsf{T}) d^Y_{i_k},
\end{equation}
where $d^Y_{i_1},\cdots,d^Y_{i_k}$ are descriptors of $\mathbf{y}_{i_1},\cdots,\mathbf{y}_{i_k}$, and $\overline{\mathbf{W}}_Q$ and $\overline{\mathbf{W}}_K$ are learnable weights.
Inspired by PointDSC~\cite{bai2021pointdsc}, we construct a compatibility graph $\mathbf{B}$ (Eq.~\ref{eq:compatibility}) of sparse keypoint correspondences $\{(\mathbf{x}_i,\hat{\mathbf{y}}_i)\}$ for spatial consistency filtering via compatibility graph embedding:
\begin{equation}
\mathbf{E}^{(l+1)}=\text{softmax}\left(\frac{1}{\sqrt{D_e}} \mathbf{E}^{(l)}\mathbf{W}_Q^{(l)}  
(\mathbf{E}^{(l)}\mathbf{W}_K^{(l)})^\mathsf{T} \odot \mathbf{B} \right)\mathbf{E}^{(l)}\mathbf{W}_V^{(l)},\; \mathbf{E}^{(0)}_i = \text{MLP}([ \mathbf{x}_i,d^X_{i}, \hat{\mathbf{y}}_i, \hat{d}^Y_{i} ]),
\label{eq:embed}
\end{equation}
where $\mathbf{E}^{(l)}$ is the correspondence-wise embedding of the $l$-th layer with learnable weights $\mathbf{W}_Q^{(l)}$, $\mathbf{W}_K^{(l)}$, $\mathbf{W}_V^{(l)}$. Finally, the embedding is fed into an MLP to classify if a correspondence is an inlier.
The predicted inlier confidences serve as the weights of keypoint correspondences for pose estimation formulated as Eq.~\ref{eq:registration}, which can be analytically solved by weighted Kabsch algorithm~\cite{kabsch1976svd}.
It is noteworthy that the above process is really lightweight and scalable to large-scale registration tasks.

After aligning two point clouds based on sparse matching, we propose to refine the transformation based on dense matching.
We still utilize local attention (Eq.~\ref{eq:local}) to predict the correspondences of dense points $\mathbf{X}^{1/2}$ from its neighbors in $\mathbf{Y}^{1/2}$ within a radius $R_d$, and we simply set the confidence weight of a correspondence with a distance $d$ as $w = [1 - d/R_d]^+$.
By solving Eq.~\ref{eq:registration} again with both sparse and dense correspondences, we can achieve more accurate pose estimation efficiently.

\subsection{Loss Functions}
Our loss function needs to supervise four modules, \textit{i.e.}, keypoint detection, coarse matching, keypoint matching, and dense registration.
For keypoint detection, we utilize the probabilistic chamfer loss~\cite{li2019usip} $\mathcal{L}_p$ to minimize the distances between the closest keypoints from the source and target point clouds after alignment under the ground-truth transformation.
Please refer to~\cite{li2019usip} for details.
\paragraph{Coarse Matching.}
Given the ground-truth coarse correspondence set $\mathcal{C}$ with an overlap ratio $o_{ij}$ for each correspondence $(i,j)\in \mathcal{C}$, we propose a spot matching loss $\mathcal{L}_s$ and a coarse matching loss $\mathcal{L}_c$ formulated as weighted cross entropy losses to supervise the layer-wise coarse matching scores $\mathbf{P}^{(l)}\;(l=1,2,\cdots,L)$ and the final coarse matching scores $\mathbf{P}$, respectively:
\begin{equation}
\mathcal{L}_s = -\frac{1}{L}\sum_{l=1}^L \frac{1}{\sum_{(i,j)\in \mathcal{C}}o_{ij}} \sum_{(i,j)\in \mathcal{C}} o_{ij}\log \mathbf{P}_{ij}^{(l)},
\end{equation}
\begin{equation}
\mathcal{L}_c = -\frac{1}{\sum_{(i,j)\in \mathcal{C}}o_{ij}} \sum_{(i,j)\in \mathcal{C}} o_{ij}\log \mathbf{P}_{ij}
-\frac{1}{|\mathcal{N}_X|}\sum_{k\in \mathcal{N}_X} \log(1-\hat{o}_k^X)
-\frac{1}{|\mathcal{N}_Y|}\sum_{k\in \mathcal{N}_Y} \log(1-\hat{o}_k^Y),
\end{equation}
where $\mathcal{N}_X$ and $\mathcal{N}_Y$ are sets of semi-dense nodes in point clouds $\mathbf{X}$ and $\mathbf{Y}$ without correspondences, respectively.
Two nodes are considered as a pair of coarse correspondence only when their ground-truth overlap ratio is greater than 0.
Assuming that the patch centered at a point $p\in\mathbb{R}^3$ is a spherical neighborhood of radius $r$, the overlapping ratio $o_{ij}$ of patches centered at $p_i\in\mathbf{X}^S$ and $q_j\in\mathbf{Y}^S$ with ground-truth translation $\mathbf{t}\in\mathbb{R}^3$ and rotation $\mathbf{R}\in SO(3)$ can be calculated by:
\begin{equation}
\label{eq:overlap}
o_{ij} = \frac{2\pi\int_{d/2}^r(r^2-h^2)dh}{4\pi r^3/3} = 1 - \frac{3d}{4r} + \frac{d^3}{16r^3}, \;d=\max\{\Vert \mathbf{R}p_i +\mathbf{t} - q_i\Vert, 2r\}.
\end{equation}

\paragraph{Keypoint Matching.}
As our keypoint matching module follows a three-stage pipeline including similarity calculation, correspondence prediction, and consistency filtering, it is reasonable to supervise these stages with three losses, respectively.
Only valid keypoint-to-patch correspondences are supervised during training, \textit{i.e.}, the distance of the keypoint $x$ and its closest point $p_x$ in the patch $C_x$ is less than a pre-defined threshold $R_p>0$, and the points whose distances from $x$ are greater than a pre-defined threshold $R_n\geq R_p$ form a non-empty set $N_x\subset C_x$.
We formulate the keypoint matching loss $\mathcal{L}_f$ as an InfoNCE loss~\cite{oord2018infonce} with symmetric learnable weights $W$, which aims at maximizing the similarity between descriptors $d_x$ and $d_{p_x}$ of true correspondences $(x,p_x)$ and minimizing the similarity between descriptors $d_x$ and $d_{n_x}$ of false correspondences $(x,n_x), n_x\in N_x$.
\begin{equation}
\label{eq:fine}
\mathcal{L}_f = -\mathbb{E}_{(x,p_x,N_x)}\left[ \log\frac{\exp(d_x^\mathsf{T}Wd_{p_x})}{\exp(d_x^\mathsf{T}Wd_{p_x}) + \sum_{n_x\in N_x}\exp(d_x^\mathsf{T}Wd_{n_x})}\right].
\end{equation}
For correspondence prediction, we adopt a $L_2$ loss $\mathcal{L}_k = \mathbb{E}_{(x,\hat{y})} \Vert \mathbf{R}x + \mathbf{t} - \hat{y} \Vert_2$ for the predicted correspondences $\hat{y}$ of keypoints $x$ from all valid keypoint-to-patch correspondences $(x,C_x)$.
For consistency filtering, we simply utilize a binary entropy loss $\mathcal{L}_i$ to supervise the confidence scores of all keypoint correspondences. The binary ground-truth label of a keypoint correspondence $(x,\hat{y})$ is $1$ if and only if it is an inlier, \textit{i.e.}, $\Vert \mathbf{R}x + \mathbf{t} - \hat{y} \Vert_2$ is less than a threshold $R_f>0$.

\paragraph{Dense Registration.} Given the translation $\hat{\mathbf{t}}$ and rotation $\hat{\mathbf{R}}$ estimated by dense registration, we adopt a translation loss $\mathcal{L}_t = \Vert \hat{\mathbf{t}} - \mathbf{t}\Vert_2$ and a rotation loss $\mathcal{L}_r = \Vert \hat{\mathbf{R}}^{\mathsf{T}}\mathbf{R} - \mathbf{I}_{3\times 3} \Vert_F$ for supervision.

Finally, we formulate our loss as $\mathcal{L}=\mathcal{L}_p + \lambda_s \mathcal{L}_s + \lambda_c\mathcal{L}_c + \lambda_f\mathcal{L}_f + \lambda_k\mathcal{L}_k + \lambda_i\mathcal{L}_i + \lambda_t \mathcal{L}_t + \lambda_r \mathcal{L}_r$, where $\lambda_c,  \lambda_s,\lambda_f, \lambda_k, \lambda_i, \lambda_t, \lambda_r$ are balancing weights.

\section{Experiments}
\label{sec:experiment}

In this section, we evaluate our method on both outdoor LiDAR point cloud datasets KITTI~\cite{geiger2012kitti}, nuScenes~\cite{nuscenes2019}, and the indoor RGBD point cloud dataset 3DMatch~\cite{zeng20173dmatch}.
Our network is trained using an AdamW~\cite{loshchilov2018adamw} optimizer with a batch size of 1, an initial learning rate of 1e-4, and a weight decay of 1e-4.
The step scheduler decrease the learning rate to 90\% every five steps, with gradients clipped at a norm of 0.5 during back propagation.
Despite the complexity of the loss function, only one stage is needed for training.
Our model is trained on an NVIDIA RTX 3090 GPU with an Intel Xeon CPU
@2.90GHZ for 5, 40, and 3 epochs on 3DMatch, KITTI, and nuScenes, respectively, and we set $\lambda_f = \lambda_i = 1, \lambda_r = 20, \lambda_t = 5, \lambda_s=0.1$, and $\lambda_c=0.2, \lambda_k=1$ for KITTI and nuScenes, $\lambda_c=1, \lambda_k=10$ for 3DMatch.

\subsection{Outdoor Scenarios: KITTI and NuScenes}
KITTI~\cite{geiger2012kitti} is a popular benchmark for autonomous driving.
Following~\cite{bai2020d3feat}, we use sequences 0 to 5 for training, 6 to 7 for validation, and 8 to 10 for testing, and select only point cloud pairs at least 10m away from each other with ICP-refined~\cite{icp1992tpami} GPS localization results as ground truth.
NuScenes~\cite{nuscenes2019} is another large-scale outdoor autonomous driving benchmark including 850 scenes for training and validation and 150 for testing. Following~\cite{lu2023hregnet}, we select each LiDAR keyframe with the second keyframe after it as a pair of point clouds.
We use three metrics for evaluation~\cite{huang2021predator}: \textit{relative translation error} (RTE), \textit{relative rotation error} (RRE), and \textit{registration recall} (RR).

\begin{table}[H]
\footnotesize
\caption{Registration performance on KITTI odometry dataset.}
\vspace{-2mm}
\label{tab:kitti}
\centering
\setlength\tabcolsep{4pt}
\begin{tabular}{lc|ccc}
\toprule
Model & Publication & RTE (cm) & RRE (°) & RR (\%) \\
\cmidrule{1-5}
3DFeat-Net & ECCV 2018~\cite{yew20183dfeat} & 25.9 & 0.25 & 96.0\\
FCGF & ICCV 2019~\cite{choy2019fcgf} & 9.5 & 0.30 & 96.6\\
D3Feat &CVPR 2020~\cite{bai2020d3feat} & 7.2 & 0.30 & 99.8\\
SpinNet &CVPR 2021~\cite{ao2021spinnet} & 9.9 & 0.47 & 99.1\\
Predator &CVPR 2021~\cite{huang2021predator} & 6.8 & 0.27 & 99.8\\
CoFiNet & NeurIPS 2021~\cite{yu2021cofinet} & 8.2 & 0.41 & 99.8\\
GeoTransformer & CVPR 2022~\cite{qin2023geotransformer} & 6.8 & 0.24 & 99.8\\
OIF-Net & NeurIPS 2022~\cite{yang2022one} & 6.5 & \textbf{0.23} & 99.8\\
PEAL & CVPR 2023~\cite{yu2023peal} & 6.8 & \textbf{0.23} & 99.8\\
DiffusionPCR & CVPR 2024~\cite{chen2023diffusionpcr} & 6.3 & \textbf{0.23} & 99.8 \\
MAC & CVPR 2023~\cite{zhang2023mac} & 8.5 & 0.40 & 99.5\\
CAST & & \textbf{2.5} & 0.27 & \textbf{100.0}\\
\bottomrule
\end{tabular}
\vspace{-3mm}
\end{table}

\begin{figure}
\centering
\includegraphics[width=\linewidth]{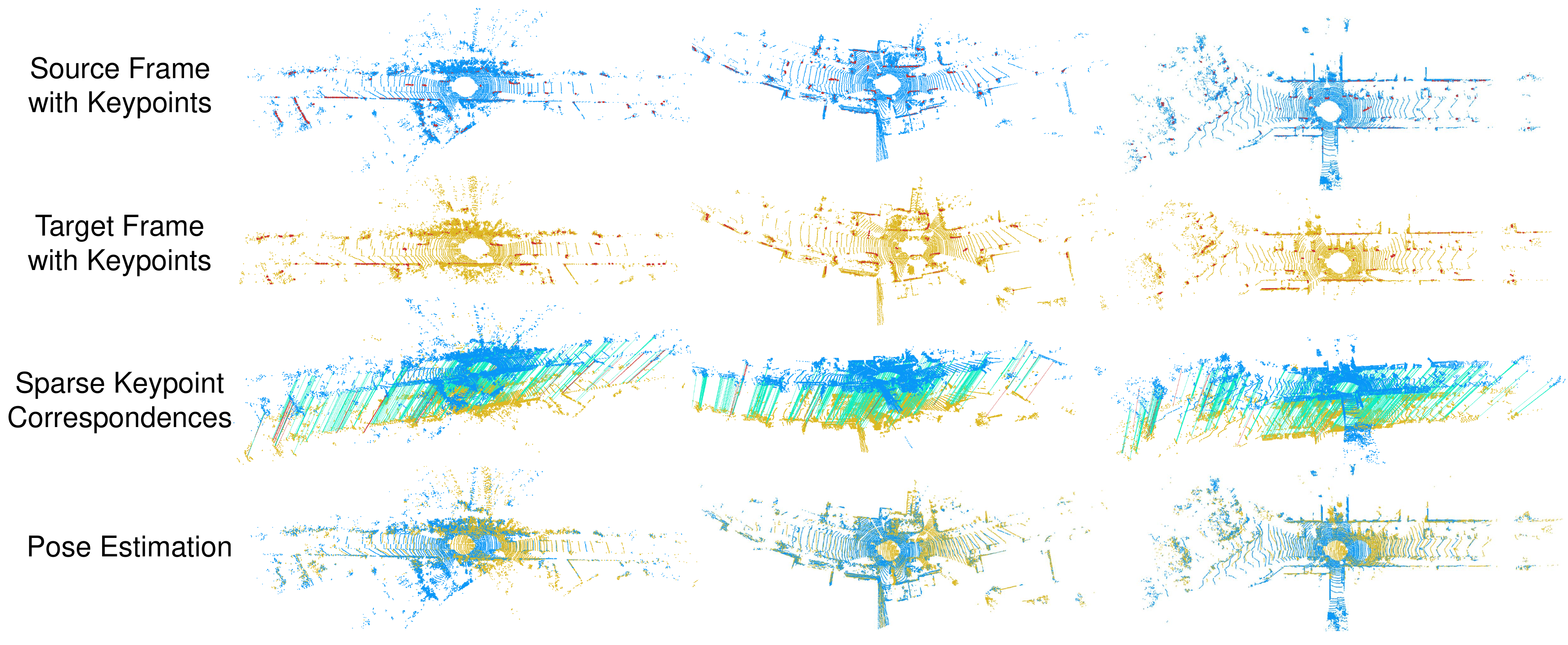}
\caption{Qualitative registration results on KITTI dataset. We show three examples in three columns. The first two rows present the raw point clouds and highlight the 3D keypoints with low uncertainty in red. Our keypoints are typically located in sharp corners and edges of buildings, pillars, and vehicles. The third row shows the predicted sparse keypoint correspondences with high scores, while the last row presents the aligned point clouds after pose estimation. Although a few outliers colored in red have not been filtered out, their distances are acceptable for accurate registration.}
\label{fig:kitti}
\vspace{-2mm}
\end{figure}

Our results on KITTI benchmark are detailed quantitatively in Table~\ref{tab:kitti} and qualitatively in Figure~\ref{fig:kitti}. Table~\ref{tab:kitti} shows that the proposed CAST outperforms various learning-based methods, including descriptor-based ~\cite{yew20183dfeat,choy2019fcgf,bai2020d3feat,ao2021spinnet,huang2021predator}, coarse-to-fine correspondence-based~\cite{yu2021cofinet,qin2023geotransformer,yang2022one} including the latest ones with iterative refinement~\cite{yu2023peal,chen2023diffusionpcr}, and a recent graph-based baseline~\cite{zhang2023mac}.
Specifically, CAST achieves an RR of 100.0\% and the lowest RTE of 2.5cm, which is 60.3\% improvement over the \textit{state-of-the-art} DiffusionPCR~\cite{chen2023diffusionpcr}, highlighting its superior robustness and accuracy.

As for the rotation error, our method slightly underperforms some coarse-to-fine methods~\cite{qin2023geotransformer, yang2022one,yu2023peal,chen2023diffusionpcr}, primarily due to numerical inaccuracies in SVD-based pose estimation that usually produces non-orthonormal rotation matrices. Typically, the rotation error is set as 0 when $\text{trace}(\hat{\mathbf{R}}^{\mathsf{T}}\mathbf{R})>3$, a condition met frequently across all methods in our tests. Consequently, RRE based on the geodesic distance may not accurately reflect the actual performance, as it is affected by these numerical errors.

\begin{wraptable}{r}{0.53\textwidth}
\centering
\footnotesize
\setlength\tabcolsep{4pt}
\caption{Registration performance on nuScenes.}
\vspace{-2mm}
\label{tab:nuscenes}
\begin{tabular}{l|ccc}
\toprule
Method & RTE (m) & RRE (°) & RR (\%) \\
\cmidrule{1-4}
Point-to-Point ICP~\cite{icp1992tpami} & 0.25 & 0.25 & 18.8\\
Point-to-Plane ICP~\cite{icp1992tpami} & 0.15 & 0.21 & 36.8\\
FGR~\cite{zhou2016fgr} & 0.71 & 1.01 & 32.2\\
RANSAC~\cite{fischler1981ransac} & 0.21 & 0.74 & 60.9\\
DCP~\cite{wang2019dcp} & 1.09 & 2.07 & 56.8\\
IDAM~\cite{li2020idam} & 0.47 & 0.79 & 88.0\\
FMR~\cite{huang2020fmr} & 0.60 & 1.61 & 92.1\\
DGR~\cite{choy2020dgr} & 0.21 & 0.48 & 98.4\\
HRegNet~\cite{lu2023hregnet} & 0.18 & 0.45 & \textbf{99.9}\\
CAST & \textbf{0.12} & \textbf{0.20} & \textbf{99.9}\\
\bottomrule
\end{tabular}
\vspace{-4mm}
\end{wraptable}
For a more challenging LiDAR benchmark nuScenes, we compare CAST with both traditional~\cite{icp1992tpami,zhou2016fgr,fischler1981ransac} and learning-based algorithms~\cite{wang2019dcp, li2020idam,huang2020fmr,choy2020dgr,lu2023hregnet} in Table~\ref{tab:nuscenes}.
We do not include the coarse-to-fine methods since none have been trained or tested on nuScenes.
Most of the results are borrowed from~\cite{lu2023hregnet} while HRegNet~\cite{lu2023hregnet} is re-evaluated with their open source codes. Our method achieves the lowest translation error of 0.12m and the lowest rotation error of 0.20° while maintaining the best RR of 99.9\%, showcasing \textit{state-of-the-art} robustness and accuracy.

\subsection{Indoor Scenarios: 3DMatch and 3DLoMatch}
Our approach is also evaluated on indoor benchmarks 3DMatch~\cite{zeng20173dmatch} and 3DLoMatch~\cite{huang2021predator}, which consist of point cloud pairs with overlaps $>$30\% and $10\%\sim30\%$, respectively.
In Table~\ref{table:3dmatch}, we use registration recall~\cite{zeng20173dmatch} as our evaluation metric, and test the runtime of all methods in Pytorch implementation with 5000 points on our device with an Intel CPU i7-12800HX@2.30GHZ and an NVIDIA RTX 3080Ti GPU for fairness, except~\cite{gojcic2019perfect} in Tensorflow implementation and~\cite{yang2022one,yu2023peal,chen2023diffusionpcr} using the results reported in their papers~\cite{yang2022one,chen2023diffusionpcr} due to the absence of source codes.
Our method along with other sparse matching baselines~\cite{yew2022regtr,ao2023buffer} directly uses all points for evaluation.
To enhance the robustness in low overlapping cases, our method is combined with RANSAC estimating an initial pose from only 250 coarse correspondences to reject the outliers during fine matching.

On the 3DMatch benchmark, our method achieves \textit{state-of-the-art} RR of 95.2\%. On the more challenging 3DLoMatch, CAST achieves a high RR of 75.1\%, outperforming all descriptors and non-iterative correspondence-based methods~\cite{yew2022regtr,yu2021cofinet,qin2023geotransformer, yu2023roitr,ao2023buffer, chen2023sira} except OIF-Net~\cite{yang2022one} using more than 1000 sampled points.
As our method typically detects about 1000 sparse keypoints and establishes less than 250 keypoint correspondences on 3DLoMatch, it is fair to compare CAST with other methods using only 250 sample points.
However, CAST outperforms the \textit{state-of-the-art} non-iterative correspondence-based methods OIF-Net~\cite{yang2022one} using less than 1000 points.
Notably, our method achieves such superior performance only with the lowest runtime, while RANSAC remains efficient due to our high inlier ratio.
Although PEAL~\cite{yu2023peal} and DiffusionPCR~\cite{chen2023diffusionpcr} show higher RR on 3DLoMatch, their iterative feature matching with overlap priors is extremely time-consuming (10 times of ours), while PEAL even requires extra information from 2D images.

\begin{table}
\footnotesize
\caption{Evaluation results on indoor RGBD point cloud datasets.}
\vspace{-2mm}
\label{table:3dmatch}
\centering
\setlength\tabcolsep{4pt}
\begin{tabular}{cc|ccccc|ccccc|c}
\toprule
& Dataset & \multicolumn{5}{c|}{3DMatch} & \multicolumn{5}{c|}{3DLoMatch} & Average\\
\cmidrule{1-13}
& & \multicolumn{5}{c|}{Registration Recall (\%)} & \multicolumn{5}{c|}{Registration Recall (\%)} & Time (s) \\
& Samples & 5000 & 2500 & 1000 & 500 & 250 & 5000 & 2500 & 1000 & 500 & 250 & All\\
\cmidrule{1-13}
\multirow{6}{*}{\rotatebox[origin=c]{90}{descriptor-based}} & PerfectMatch~\cite{gojcic2019perfect}
& 78.4 & 76.2 & 71.4 & 67.6 & 50.8
& 33.0 & 29.0 & 23.3 & 17.0 & 11.0&-\\
& FCGF~\cite{choy2019fcgf}
& 85.1 & 84.7 & 83.3 & 81.6 & 71.4
& 40.1 & 41.7 & 38.2 & 35.4 & 26.8 & 0.271\\
& D3Feat~\cite{bai2020d3feat}
& 81.6 & 84.5 & 83.4 & 82.4 & 77.9
& 37.2 & 42.7 & 46.9 & 43.8 & 39.1 &0.289\\
& SpinNet~\cite{ao2021spinnet}
& 88.6 & 86.6 & 85.5 & 83.5 & 70.2
& 59.8 & 54.9 & 48.3 & 39.8 & 26.8 & 90.804\\
& YOHO~\cite{wang2022yoho}
& 90.8 & 90.3 & 89.1 & 88.6 & 84.5
& 65.2 & 65.5 & 63.2 & 56.5 & 48.0 & 13.529\\
& Predator~\cite{huang2021predator}
& 89.0 & 89.9 & 90.6 & 88.5 & 86.6
& 59.8 & 61.2 & 62.4 & 60.8 & 58.1 & 0.759\\
\cmidrule{1-13}
\multirow{10}{*}{\rotatebox[origin=c]{90}{correspondence-based}}
& REGTR~\cite{yew2022regtr} & & & 92.0 & &
& & & 64.8 & & & 0.382\\
& CoFiNet~\cite{yu2021cofinet}
& 89.3 & 88.9 & 88.4 & 87.4 & 87.0
& 67.5 & 66.2 & 64.2 & 63.1 & 61.0 &0.306\\
& GeoTransformer~\cite{qin2023geotransformer}
& 92.0 & 91.8 & 91.8 & 91.4 & 91.2
& 75.0 & 74.8 & 74.2 & 74.1 & 73.5 & 0.192\\
& OIF-Net~\cite{yang2022one}
& 92.4 & 91.9 & 91.8 & 92.1 & 91.2 
& 76.1 & 75.4 & 75.1 & 74.4 & 73.6 & 0.555\\
& RoITr~\cite{yu2023roitr}
& 91.9 & 91.7 & 91.8 & 91.4 & 91.0 
& 74.7 & 74.8 & 74.8 & 74.2 & 73.6 & 0.457\\
& PEAL~\cite{yu2023peal}
& 94.4 & 94.1 & 94.1 & 93.9 & 93.4 
& 79.2 & 79.0 & 78.8 & 78.5 & 77.9 & 2.074\\
& BUFFER~\cite{ao2023buffer} & & & 92.9 & & 
& & & 71.8 & & & 0.290 \\
& SIRA-PCR~\cite{chen2023sira}
& 93.6 & 93.9 & 93.9 & 92.7 & 92.4 
& 73.5 & 73.9 & 73.0 & 73.4 & 71.1 & 0.291\\
& DiffusionPCR~\cite{chen2023diffusionpcr}
& 94.4 & 94.3 & 94.5 & 94.0 & 93.9 
& \textbf{80.0} & \textbf{80.4} & \textbf{79.2} & \textbf{78.8} & \textbf{78.8} & 1.964\\
& CAST & & & \textbf{95.2} & &  & & & 75.1 & & & \textbf{0.182} \\
\bottomrule
\end{tabular}
\vspace{-2mm}
\end{table}

\subsection{Ablation Studies}
We select indoor datasets for ablation studies of coarse matching as they are more challenging.
Here we evaluate the RR over the whole dataset rather than the average RR of eight sequences reported in Table~\ref{table:3dmatch}, which is more reasonable for a dataset with significant variances of sequence lengths. Besides, we assess two extra metrics to directly measure the performance of coarse matching: \textit{patch inlier ratio} (PIR), the fraction of patch matches with actual overlap; and \textit{patch matching recall} (PMR), the fraction of point cloud pairs with PIR above 20\%.
Results from five experiments in Table~\ref{tab:coarse} demonstrate the effects of the proposed multi-scale feature fusion (MS), spot-guided cross-attention (SG), consistency-aware self-attention (CA), and the overlap head for overlap score prediction (OV).
The first experiment ablating CA and replacing SG with linear cross-attention, suffers performance degradation in all metrics due to inconsistency.
The second experiment improves all metrics based on SG, while the last one achieves the best performance via CA, showcasing their effectiveness. Figure~\ref{fig:attention} visualizes the vanilla global cross-attention and our spot-guided cross-attention. Instead of interacting with many similar yet irrelevant regions for misleading feature aggregation, SG can effectively select instructive areas to attend to according to local consistency.
Compared to the last experiment, the third one verifies the effectiveness of multi-scale feature fusion, while the fourth one demonstrates the necessity of overlap prediction.

\begin{table}
\footnotesize
\caption{Ablation studies of coarse matching modules on indoor datasets.}
\vspace{-2mm}
\label{tab:coarse}
\centering
\setlength\tabcolsep{4pt}
\begin{tabular}{c|cccc|ccc|ccc}
\toprule
& \multirow{2}{*}{MS} & \multirow{2}{*}{SG} & \multirow{2}{*}{CA} &
\multirow{2}{*}{OV} & \multicolumn{3}{c|}{3DMatch} & \multicolumn{3}{c}{3DLoMatch}\\
& & & & & PIR (\%) & PMR (\%) & RR (\%) & PIR (\%) & PMR (\%) & RR (\%)\\
\cmidrule{1-11}
1 & \checkmark & & & \checkmark & 77.56 & 95.87 & 94.45 & 40.82 & 70.58 & 72.07\\
2 & \checkmark & \checkmark & & \checkmark & 77.95 & 96.61 & 94.92 & 42.55 & 72.77 & 74.57\\
3 & & \checkmark & \checkmark & \checkmark & 69.58 & 96.67 & 94.14 & 32.59 & 65.02 & 73.00\\
4 & \checkmark & \checkmark & \checkmark &  & 73.56 & \textbf{97.17} & 95.07 & 35.25 & 68.33 & 74.91 \\
5 & \checkmark & \checkmark & \checkmark & \checkmark & \textbf{79.79} & \textbf{97.17} & \textbf{96.01} & \textbf{44.41} & \textbf{75.24} & \textbf{76.59}\\
\bottomrule
\end{tabular}
\end{table}

\begin{wraptable}{r}{0.59\textwidth}
\footnotesize
\caption{Ablation studies of fine matching on KITTI.}
\vspace{-1mm}
\label{tab:fine}
\centering
\setlength\tabcolsep{4pt}
\begin{tabular}{l|ccc}
\toprule
& RTE (cm) & RRE (°) & RR (\%) \\
\cmidrule{1-4}
ours & \textbf{2.51} & \textbf{0.27} & \textbf{100.00} \\
ours w/o dense registration & 3.13 & 0.30  & \textbf{100.00}\\
ours w/o virtual dense corr. & 2.85 & 0.28  & \textbf{100.00}\\
ours w/o keypoint detection & 3.58 & 0.30  & \textbf{100.00}\\
ours w/o virtual sparse corr. & 3.25 & 0.30  & \textbf{100.00}\\
ours w/o graph embedding & 5.01 & 0.30  & \textbf{100.00}\\
\bottomrule
\end{tabular}
\vspace{-3mm}
\end{wraptable}
Additionally, we conducted five ablation studies on KITTI for a better understanding of our fine matching, since pose errors are better metrics to reflect accuracy.
The second experiment using only sparse keypoints for registration highlights the effectiveness of dense registration,
while the third one shows the effect of learnable dense correspondences compared to nearest neighbors.
The last three experiments report the performance of sparse registration by ablating the keypoint detector, the learnable sparse correspondences, and the compatibility graph embedding, each demonstrating their necessity for accuracy.
Despite these variations, all studies maintain a 100\% RR, showing the robustness of coarse matching.

\section{Conclusion}
In this paper, we present a novel consistency-aware spot-guided Transformer to achieve compact and consistent coarse matching for point cloud registration.
At the coarse matching stage, our consistency-aware self-attention enhances the feature representations with sparse sampling from the geometric compatibility graph. Additionally, our spot-guided cross-attention leverages local consistency to guide the cross-attention to confident spots without interfering with irrelevant areas.
Based on these semi-dense and consistent coarse correspondences, a lightweight and scalable sparse-to-dense fine matching module empowered by local attention can achieve accurate pose estimation without optimal transport or hypothesis-and-selection pipelines.
Our method has showcased \textit{state-of-the-art} accuracy, robustness, and efficiency for point cloud registration across different 3D sensors and scenarios, which paves the way for large-scale real-time applications such as SLAM.

\section*{Acknowledgments}
This work is supported by the National Natural Science Foundation of China (62203383/62088101).

{\small
\bibliographystyle{unsrt}
\bibliography{cast}
}

\newpage

\appendix

\section{Appendix}
\label{sec:appendix}
In this appendix, we will first detail our neural network architecture with hyper-parameters in Sec.~\ref{subsec:architecture}. Then we introduce the related datasets in Sec.~\ref{subsec:data} and provide more quantitative experimental results of our method with detailed explanation of evaluation metrics in Sec.~\ref{subsec:metric}. Additionally, more qualitative results are illustrated in Sec.~\ref{subsec:qualitative}. Finally, we will discuss the limitations and broader impacts in Sec.~\ref{subsec:limitation}
and Sec.~\ref{subsec:impact}, respectively.

\subsection{Neural Network Architectures and Hyper-parameters}
\label{subsec:architecture}
\begin{figure}[b]
\centering
\includegraphics[width=\linewidth]{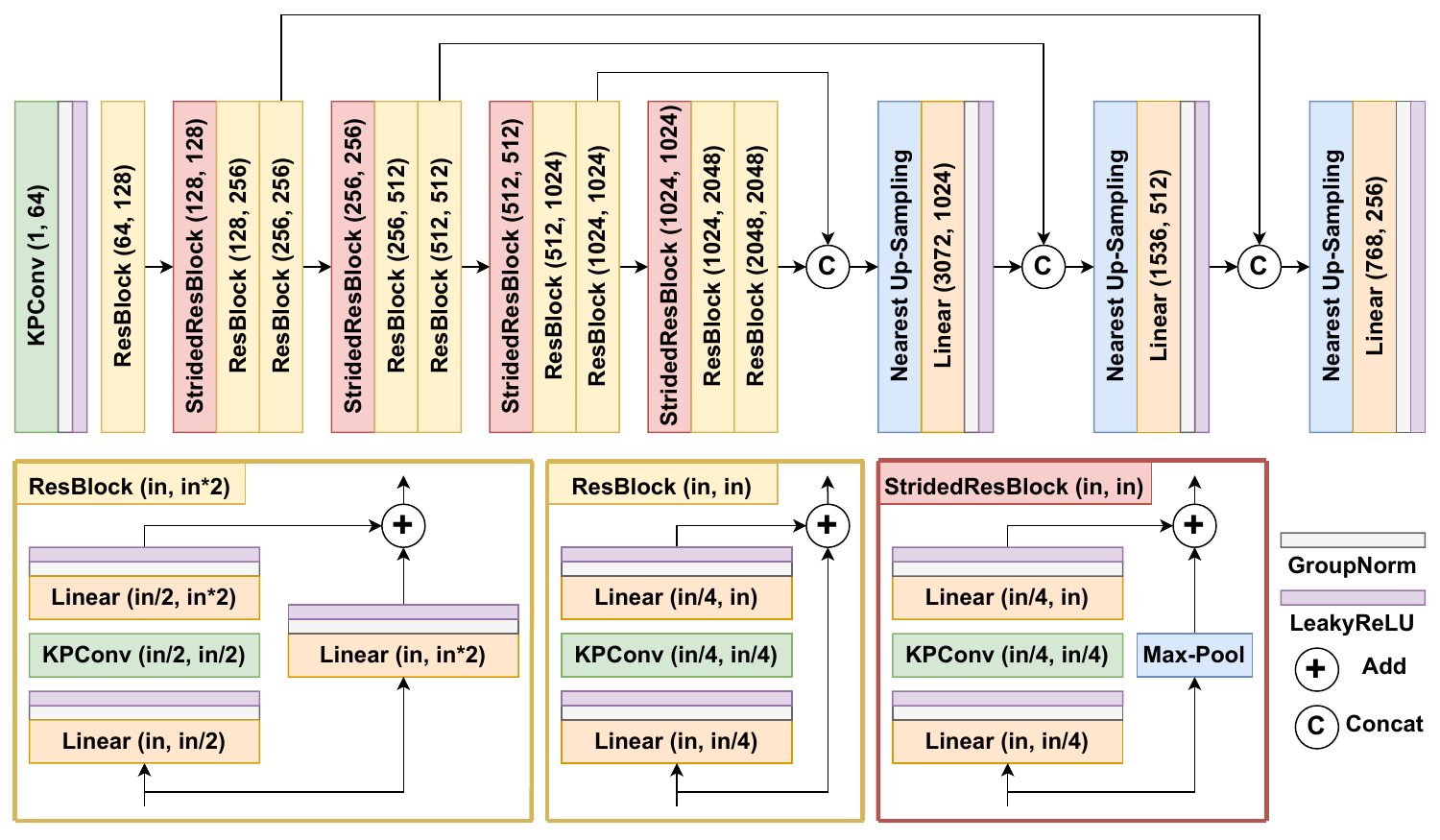}
\caption{The detailed architecture of the KPConv-based feature pyramid network.}
\label{fig:backbone}
\end{figure}

\paragraph{Feature Pyramid Network.}
CAST utilizes a fully convolutional feature pyramid network as the backbone, which follows an encoder-decoder architecture based on KPConv~\cite{thomas2019kpconv} operations.
Details of our network
architecture are illustrated in Figure~\ref{fig:backbone}, which remain the same as~\cite{qin2023geotransformer} including five encoder layers and three decoder layers.
Note that the backbone for indoor datasets 3DMatch~\cite{zeng20173dmatch} is slightly different, which only comprises four encoder layers and two decoder layers.

\begin{figure}
\centering
\includegraphics[width=0.85\linewidth]{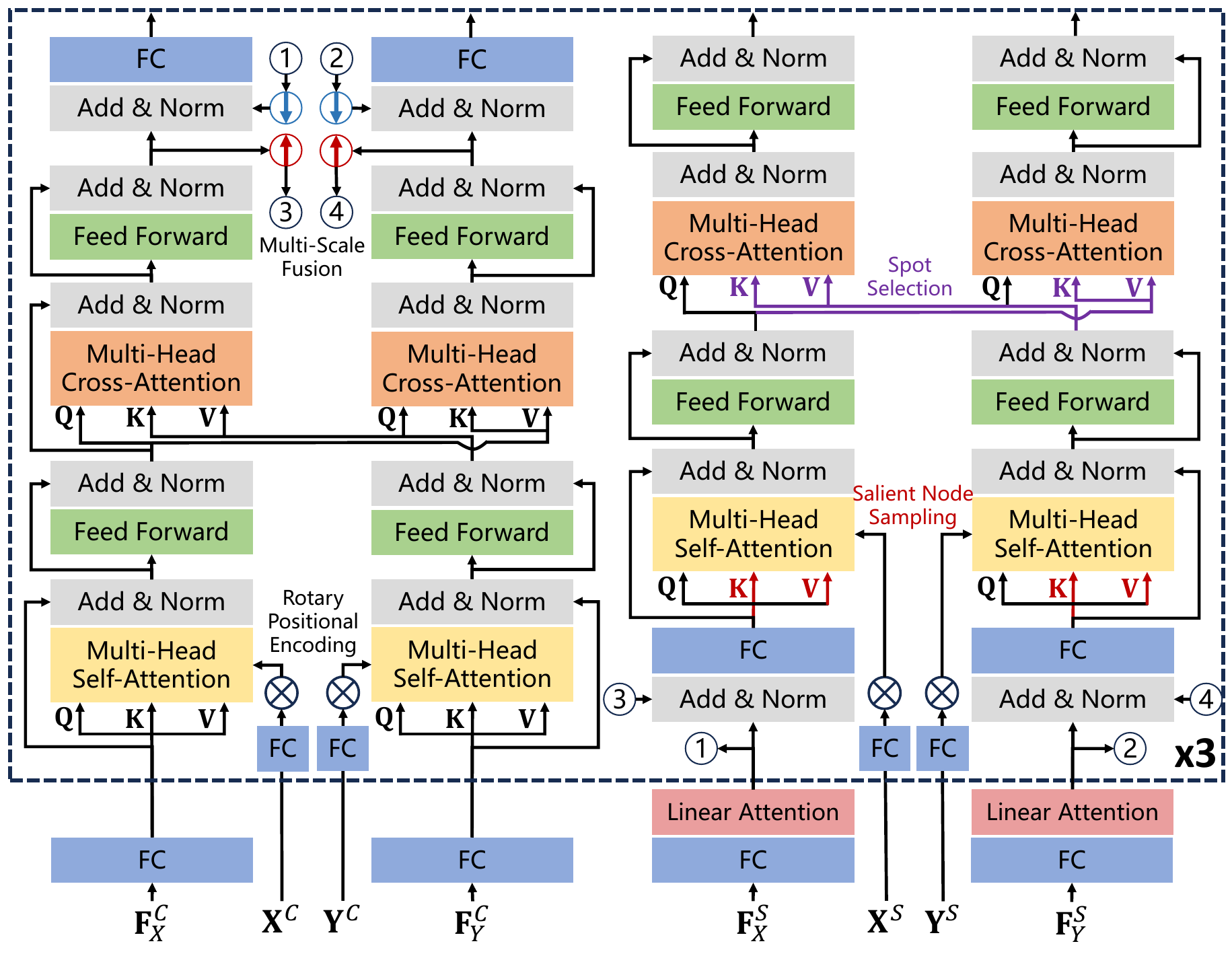}
\caption{The detailed architecture of the consistency-aware spot-guided Transformer with multi-scale feature fusion for coarse feature matching.}
\label{fig:transformer}
\end{figure}

\paragraph{Coarse Matching Module.}
Figure~\ref{fig:transformer} illustrates the details of our consistency-aware spot-guided attention blocks for coarse matching.
Both coarse features and semi-dense features extracted from the backbone are first projected to 128 dimensions and then pass through three consistency-aware spot-guided attention blocks (Figure~\ref{fig:overview}).
Each attention module uses 4 heads with ReLU activation.
Compared to vanilla attention with a quadratic increase in the size of the attention matrix with respect to the input length, the linear attention~\cite{katharopoulos2020transformers} is much more efficient in global context aggregation by replacing the softmax operator with the product of two kernel functions:
\begin{equation}
\text{Linear attention}(Q,K,V)= \phi(Q) (\phi(K)^\mathsf{T} V),
\end{equation}
where $\phi(\cdot)=\text{elu}(\cdot)+1$.
For spot-guided cross-attention, each node $\mathbf{x}_i^S$ selects 4 seeds as $\mathcal{N}_s(\mathbf{x}_i^S)$ from the neighborhood $\mathcal{N}(\mathbf{x}_i^S)$ with 12 nodes based on consistency-aware matching confidence.
For consistency-aware self-attention, we first scale the generalized degrees to $[0,1]$, and sample 48 nodes with highest matching scores from nodes with scaled degrees greater than 0.3 as keys to be attended.

\begin{wrapfigure}{r}{0.65\textwidth}
\centering
\includegraphics[width=\linewidth]{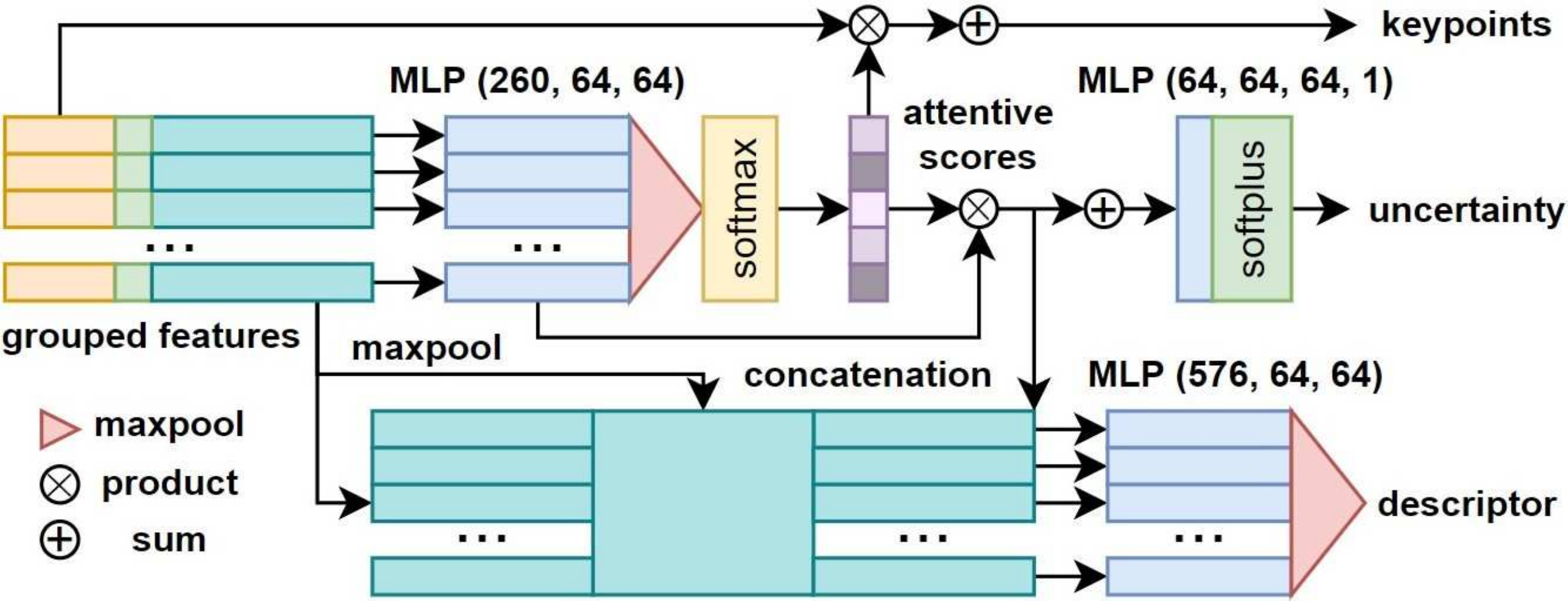}
\caption{Attentive keypoint detector and descriptor.}
\label{fig:detector}
\vspace{-2mm}
\end{wrapfigure}

\paragraph{Fine Matching Modules.}
For keypoint detection and description, we utilize an attentive keypoint detector~\cite{lu2020rskdd} to extract a keypoint with a descriptor from each local patch, which contains $k=32$ nearest neighbors among dense points $\mathbf{x}_{i_1},\cdots, \mathbf{x}_{i_k}\in\mathbf{X}^{1/2}$ with features $f_{i_1},\cdots,f_{i_k}$ of a patch node $\mathbf{x}^S_i\in \mathbf{X}^S$. The details of the keypoint detector and descriptor are illustrated in Figure~\ref{fig:detector}, which are based on only shared MLPs with $\left[\mathbf{x}_{i_j}, \Vert \mathbf{x}_{i_j} - \mathbf{x}_i^S\Vert_2, f_{i_j}\right], j=1,\cdots, k$ as inputs.
For keypoints $\mathbf{x}_1^K,\cdots,\mathbf{x}_{M'}^K$ with uncertainties $\sigma_1,\cdots,\sigma_{M'}$ and $\mathbf{y}_1^K,\cdots,\mathbf{y}_{N'}^K$ with uncertainties $\sigma_1,\cdots,\sigma_{N'}$ from point cloud $\mathbf{X}$ and $\mathbf{Y}$, respectively, we adopt the probabilistic chamfer loss in~\cite{li2019usip} for training, please refer to their paper for details.

To establish keypoint-to-patch correspondences based on coarse correspondences, each keypoint is assigned to its nearest node when their distance is below a threshold $R_{k}$. Then the keypoint corresponds to the corresponding patch of this node containing $k_p$ keypoints. As mentioned in Sec.~\ref{sec:method}, we leverage a single-head attention layer to predict the virtual correspondence of each keypoint based on only $k_s$ keypoints in its corresponding patch with the highest similarity (Eq.~\ref{eq:local}). Finally, spatial consistency filtering is performed via three graph compatibility embedding layers with embedding dimension $D_e=64$ in Eq.~\ref{eq:embed}.
After sparse matching and registration, we further refine the transformation based on dense matching, which searches $k_d$ nearest neighbors within a distance threshold $R_d$ for each dense point to predict its virtual dense correspondence.
Specifically, we set $k_s=4,k_d=6$ in our implementation, while other related hyper-parameters are listed in Table~\ref{tab:param}.

\begin{table}
\footnotesize
\centering
\begin{tabular}{c|c|ccc}
\toprule
Hyper-parameters & Explanation & 3DMatch & KITTI & nuScenes\\
\cmidrule{1-5}
$r$ & coarse overlap radius (Eq.~\ref{eq:overlap}) & 0.075m & 1.2m & 1.2m\\
$\sigma_c$ & coarse compatibility threshold (Eq.~\ref{eq:compatibility}) & 0.15m & 1.8m & 1.8m\\
$R_k$ & keypoint-to-node distance threshold & 0.1m & 1.8m & 1.8m\\
$\sigma_d$ & fine compatibility threshold & 0.1m & 1.0m & 1.0m\\
$R_d$ & dense matching radius & 0.15m & 0.75m & 1.0m\\
$R_p$ & positive matching threshold (Eq.~\ref{eq:fine}) & 0.05m & 0.45m & 0.45m\\
$R_n$ & negative matching threshold (Eq.~\ref{eq:fine}) & 0.06m & 0.6m & 0.6m\\
$k_p$ & number of keypoints in a patch & 16 & 24 & 24\\
\bottomrule
\end{tabular}
\caption{Hyper-parameter setting for three datasets in our implementation.}
\label{tab:param}
\vspace{-1mm}
\end{table}

\paragraph{Weighted Kabsch Algorithm.}
The weighted Kabsch algorithm~\cite{kabsch1976svd}, also known as the Proscrutes algorithm, provides the closed-form solution for the point cloud registration problem (Eq.~\ref{eq:registration}).
Given a predicted correspondence set $\hat{\mathcal{C}}=\{(\mathbf{x}_k,\mathbf{y}_k):k=1,\cdots,n\}$, the optimal rigid transformation $\hat{\mathbf{R}}\in SO(3), \hat{\mathbf{t}}\in \mathbb{R}^3$ can be estimated via two steps:

\noindent\textbf{Step 1.} Centralize the point clouds by subtracting away their weighted centroids:
\begin{equation}
\Tilde{\mathbf{x}}_k = \mathbf{x}_k - \overline{\mathbf{x}},\;
\Tilde{\mathbf{y}}_k = \mathbf{y}_k - \overline{\mathbf{y}},\;
\overline{\mathbf{x}} = \frac{\sum_{k=1}^n w_k \mathbf{x}_k}{\sum_{k=1}^n w_k},\;
\overline{\mathbf{y}} = \frac{\sum_{k=1}^n w_k \mathbf{y}_k}{\sum_{k=1}^n w_k}.
\end{equation}
\noindent\textbf{Step 2.} Estimate the transformation. A $3\times 3$ weighted covariance matrix can be computed as
\begin{equation}
\mathbf{H} = \sum_{k=1}^n w_k \Tilde{\mathbf{x}}_k \Tilde{\mathbf{y}}_k^\mathsf{T}.
\end{equation}
With its singular value decomposition $\mathbf{H} = \mathbf{U} \mathbf{\Sigma} \mathbf{V}^\mathsf{T}$, the optimal estimate of pose is given by:
\begin{equation}
\hat{\mathbf{R}} = \mathbf{V}\begin{bmatrix}
    1 & 0 & 0 \\  0 & 1 & 0 \\
    0 & 0 & \det(\mathbf{V}\mathbf{U}^\mathsf{T})
\end{bmatrix}\mathbf{U}^\mathsf{T},\;
\hat{\mathbf{t}} = \overline{\mathbf{y}} - \hat{\mathbf{R}}\overline{\mathbf{x}}.
\end{equation}

\subsection{Data and Benchmarks}
\label{subsec:data}
\paragraph{3DMatch.} 3DMatch~\cite{zeng20173dmatch} is a widely used RGBD point cloud dataset for 3D reconstruction, including 62 scenes from 7-Scenes~\cite{shotton2013scene7}, RGB-D Scenes V2~\cite{lai2014rgbdv2}, Analysis-by-Synthesis~\cite{valentin2016analysis}, BundleFusion~\cite{dai2017bundlefusion}, SUN3D~\cite{xiao2013sun3d} and Halber \textit{et al.}~\cite{halber2017fine} with their licenses in Table~\ref{tab:license}, where 46 scenes are used for training, 8 scenes for validation and 8 scenes for testing.
Input point clouds are generated by fusing 50 consecutive depth frames collected by RGBD cameras using TSDF volumetric fusion~\cite{curless1996volumetric}.
Different from the original 3DMatch~\cite{zeng20173dmatch} that only consists of point cloud pairs with >30\% overlaps, ~\cite{huang2021predator} also includes point cloud pairs with overlaps between 10\% and 30\% for training, and it sets two benchmarks for performance evaluation, \textit{i.e.}, 3DMatch consisting of point cloud pairs with >30\% overlaps, and 3DLoMatch consisting of point cloud pairs with low overlap ratios between 10\% and 30\%.
Following~\cite{qin2023geotransformer}, we utilize the voxel grid down-sampling of 2.5cm voxel size for data preprocessing, which picks the centroid as a down-sampled point when multiple points fall into a common voxel grid.
We leverage the data and evaluation protocols in~\cite{huang2021predator} for training and testing.

\begin{table}[H]
\footnotesize
\centering
\begin{tabular}{c|c}
\toprule
Datasets & License\\
\cmidrule{1-2}
7-Scenes~\cite{shotton2013scene7} & Non-commercial use only\\
Analysis-by-Synthesis~\cite{valentin2016analysis} & CC BY-NC-SA 4.0\\
BundleFusion~\cite{dai2017bundlefusion} & CC BY-NC-SA 4.0\\
RGB-D Scenes v2~\cite{lai2014rgbdv2} & (License not stated)\\
SUN3D~\cite{xiao2013sun3d} & CC BY-NC-SA 4.0\\
Halber \textit{et al.}~\cite{halber2017fine} & CC BY-NC-SA 4.0\\
\bottomrule
\end{tabular}
\caption{Raw data in the 3DMatch~\cite{zeng20173dmatch} dataset and their
licenses.}
\label{tab:license}
\vspace{-1mm}
\end{table}

\paragraph{KITTI.} KITTI~\cite{geiger2012kitti} is a classic benchmark under the NonCommercial-ShareAlike 3.0 License for a variety of computer vision tasks of autonomous driving, ranging from LiDAR-based or vision-based or multi-sensors based odometry, object detection and tracking, optical flow estimation, point cloud registration, \textit{etc}.
KITTI comprises of 11 sequences scanned by a Velodyne HDL-64 3D laser scanner in driving scenarios.
Following the data splitting method in~\cite{choy2019fcgf}, we use sequences 0 to 5 for training, 6 to 7 for validation, and 8 to 10 for testing. 
Besides, we directly leverage the source code of~\cite{bai2020d3feat} to select point cloud pairs which are at least 10m away from each other, which leads to 1,358 training pairs, 180 validation pairs, and 555 testing pairs.
Moreover, as the ground truth transformations provided by GPS are less accurate, we follow~\cite{bai2020d3feat} to refine them via standard ICP~\cite{icp1992tpami} in 500 iterations.
Following~\cite{bai2020d3feat}, we utilize the voxel down-sampling of 0.3m voxel size for data preprocessing.

\paragraph{NuScenes.}
NuScenes~\cite{nuscenes2019} is an outdoor autonomous driving datasets under CC BY-NC-SA 4.0 license.
It is the first large-scale dataset to provide data from the entire sensor suite of an autonomous vehicle, consisting of 850 scenes for training and validation and 150 scenes for testing.
Following~\cite{lu2023hregnet}, we select the first 700 scenes from the 850 scenes for training and the others for validation.
The information about the point cloud pairs with ground-truth transformations is downloaded from the source codes of~\cite{lu2023hregnet} which selects each LiDAR keyframe with the second keyframe after it as a pair.
For data preprocessing, we apply 0.3m voxel grid down-sampling.

\subsection{Evaluation Metrics with Extra Quantitative Results}
\label{subsec:metric}
The outdoor datasets KITTI~\cite{geiger2012kitti} and nuScenes~\cite{nuscenes2019} commonly use three metrics for evaluation~\cite{huang2021predator}: (1) \textit{relative translation error} (RTE), the Euclidean distance between the estimated and ground-truth translation vectors;
(2) \textit{relative rotation error} (RRE), the geodesic distance between the estimated and ground-truth rotation matrices on $SO(3)$;
and (3) \textit{registration recall} (RR), the percentage of point cloud pairs with RTE$<$2m and RRE$<$5°. Given the ground-truth rotation $\mathbf{R}\in SO(3)$ and translation $\mathbf{t}\in\mathbb{R}^3$, as well as the estimated rotation $\hat{\mathbf{R}}$ and translation $\hat{\mathbf{t}}$, the RRE and RTE are formulated as
\begin{equation}
\text{RRE} = \arccos{\left( \frac{\text{trace}(\hat{\mathbf{R}}^{\mathsf{T}}\mathbf{R}) - 1}{2}\right)},\;
\text{RTE} = \Vert \hat{\mathbf{t}} - \mathbf{t}\Vert_2.
\end{equation}

The indoor benchmarks 3DMatch~\cite{zeng20173dmatch} and 3DLoMatch~\cite{huang2021predator} commonly leverage three metrics for evaluation: \textit{registration recall} (RR), \textit{inlier ratio} (IR), and \textit{feature matching recall} (FMR). To keep with existing methods~\cite{huang2021predator}, we exclude the consecutive point clouds when evaluating the RR.

\paragraph{Registration Recall} refers to the percentage of point cloud pairs whose root mean square error (RMSE) is less than a pre-defined threshold $\tau=0.2$m. Given point clouds $\mathbf{X},\mathbf{Y}$ with a ground-truth correspondence set $\mathbf{C}=\{(\mathbf{x}_i,\mathbf{y}_j): \mathbf{x}_i\in\mathbf{X}, \mathbf{y}_j\in\mathbf{Y}\}$, estimated rotation $\hat{\mathbf{R}}\in SO(3)$, and estimated translation $\hat{\mathbf{t}}\in\mathbb{R}^3$, the RMSE of $\mathbf{X},\mathbf{Y}$ is formulated as
\begin{equation}
\text{RMSE}(\mathbf{X},\mathbf{Y}) = \sqrt{\frac{1}{\left| \mathbf{C} \right|}\sum_{(\mathbf{x}_i,\mathbf{y}_j) \in \mathbf{C}} \Vert \hat{\mathbf{R}}\mathbf{x}_i + \hat{\mathbf{t}} - \mathbf{y}_j \Vert^2}.
\end{equation}

\paragraph{Inlier Ratio} refers to the percentage of estimated correspondences whose distance is less than a pre-defined threshold $\tau_1=0.1$m. Given point clouds $\mathbf{X},\mathbf{Y}$ with an estimated correspondence set $\hat{\mathbf{C}}=\{(\mathbf{x}_i,\mathbf{y}_j): \mathbf{x}_i\in\mathbf{X}, \mathbf{y}_j\in\mathbf{Y}\}$, ground-truth rotation $\mathbf{R}\in SO(3)$, and ground-truth translation $\mathbf{t}\in\mathbb{R}^3$, the IR of $\mathbf{X},\mathbf{Y}$ is formulated as
\begin{equation}
\text{IR}(\mathbf{X},\mathbf{Y}) = \frac{1}{| \hat{\mathbf{C}}|}\sum_{(\mathbf{x}_i,\mathbf{y}_j) \in \hat{\mathbf{C}}} \mathds{1} \left( \Vert \mathbf{R}\mathbf{x}_i + \mathbf{t} - \mathbf{y}_j \Vert <\tau_1 \right).
\end{equation}

\paragraph{Feature Matching Recall}
refers to the fraction of point cloud pairs with an IR$>\tau_2=5$\%.

We adopt RR for evaluation as it directly measures the performance on the target task of point cloud registration, while FMR and IR are not suitable to quantize our matching performance for three reasons:
(1) our method detects only sparse keypoints for fine matching, hence it is not fair to be compared with other methods based on dense descriptors or correspondences when using the same number of sampled points for evaluation;
(2) our method directly supervises the differential pose estimator to learn virtual correspondences with confidence scores instead of explicit feature matching and outlier rejection, hence our method inevitably retains many outliers which are acceptable for registration but not likely to maintain advanced matching performance;
(3) feature matching recall and inlier ratio only reflect the matching performance which is not decisive for either accuracy (relative pose errors) or robustness (registration recall) especially for RANSAC-free methods~\cite{qin2023geotransformer, yang2022one, yu2023roitr, chen2023sira}.

\paragraph{Feature Matching Performance.}
Nevertheless, this appendix presents IR and FMR in Table~\ref{tab:big} to demonstrate the feature matching performance of our method. For IR evaluation, we scale the inlier confidences predicted by compatibility graph embedding to $[0,1]$ and discard the correspondences with confidences $<0.1$. As the number of keypoint correspondences after filtering is always less than 1000, we only report the IR regarding $\leq$1000 correspondences.
Even without filtering, the number of keypoint correspondences is always less than 2500, hence we only report the FMR regarding $\leq$2500 correspondences.
Enjoying the merits of our coarse matching, keypoint detection, and compatibility graph embedding, our method achieves the highest inlier ratio compared to all sorts of baselines.
As for FMR, our method performs on par with DiffusionPCR~\cite{chen2023diffusionpcr} on 3DMatch and better than CoFiNet~\cite{yu2021cofinet} on 3DLoMatch.
However, our method performs worse than other coarse-to-fine methods~\cite{qin2023geotransformer, yang2022one, yu2023roitr, yu2023peal, chen2023sira, chen2023diffusionpcr} on 3DLoMatch, since it remains challenging to extract enough keypoint correspondences in extremely low overlapping cases due to sparsity.

\begin{table}
\footnotesize
\caption{Inlier ratios and feature matching recalls on indoor datasets 3DMatch.}
\vspace{-1mm}
\label{tab:big}
\centering
\setlength\tabcolsep{4pt}
\begin{tabular}{cc|ccccc|ccccc}
\toprule
\multicolumn{12}{c}{Inlier Ratio (\%)}\\
\cmidrule{1-12}
& Benchmark & \multicolumn{5}{c|}{3DMatch} & \multicolumn{5}{c}{3DLoMatch} \\
& Samples & 5000 & 2500 & 1000 & 500 & 250 & 5000 & 2500 & 1000 & 500 & 250 \\
\cmidrule{1-12}
\multirow{6}{*}{\rotatebox[origin=c]{90}{descriptor-based}} & PerfectMatch~\cite{gojcic2019perfect}
& 36.0 & 32.5 & 26.4 & 21.5 & 16.4
& 11.4 & 10.1 & 8.0 & 6.4 & 4.8\\
& FCGF~\cite{choy2019fcgf}
& 56.8 & 54.1 & 48.7 & 42.5 & 34.1
& 21.4 & 20.0 & 17.2 & 14.8 & 11.6\\
& D3Feat~\cite{bai2020d3feat}
& 39.0 & 38.8 & 40.4 & 41.5 & 41.8
& 13.2 & 13.1 & 14.0 & 14.6 & 15.0\\
& SpinNet~\cite{ao2021spinnet}
& 47.5 & 44.7 & 39.4 & 33.9 & 27.6
& 20.5 & 19.0 & 16.3 & 13.8 & 11.1\\
& YOHO~\cite{wang2022yoho}
& 64.4 & 60.7 & 55.7 & 46.4 & 41.2
& 25.9 & 23.3 & 22.6 & 18.2 & 15.0\\
& Predator~\cite{huang2021predator}
& 58.0 & 58.4 & 57.1 & 54.1 & 49.3
& 26.7 & 28.1 & 28.3 & 27.5 & 25.8\\
\cmidrule{1-12}
\multirow{8}{*}{\rotatebox[origin=c]{90}{correspondence-based}}
& CoFiNet~\cite{yu2021cofinet}
& 49.8 & 51.2 & 51.9 & 52.2 & 52.2
& 24.4 & 25.9 & 26.7 & 26.8 & 26.9\\
& GeoTransformer~\cite{qin2023geotransformer}
& 71.9 & 75.2 & 76.0 & 82.2 & 85.1
& 43.5 & 45.3 & 46.2 & 52.9 & 57.7\\
& OIF-Net~\cite{yang2022one}
& 62.3 & 65.2 & 66.8 & 67.1 & 67.5
& 27.5 & 30.0 & 31.2 & 32.6 & 33.1\\
& RoITr~\cite{yu2023roitr}
& \textbf{82.6} & \textbf{82.8} & 83.0 & 83.0 & 83.0
& \textbf{54.3} & \textbf{54.6} & 55.1 & 55.2 & 55.3\\
& PEAL~\cite{yu2023peal}
& 74.8 & 81.3 & 86.0 & 87.9 & 89.2
& 49.1 & 54.1 & 60.5 & 63.6 & 65.0\\
& SIRA-PCR~\cite{chen2023sira}
& 70.8 & 78.3 & 83.7 & 85.9 & 87.4
& 43.3 & 49.0 & 55.9 & 58.8 & 60.7\\
& DiffusionPCR~\cite{chen2023diffusionpcr}
& 75.0 & 81.6 & 86.3 & 88.2 & 89.4
& 49.7 & 55.4 & 61.8 & 64.5 & 66.2\\
& CAST
& - & - & \textbf{91.2} & \textbf{91.5} & \textbf{93.1}
& - & - & \textbf{66.3} & \textbf{66.3} & \textbf{66.5}\\
\cmidrule{1-12}
\multicolumn{12}{c}{Feature Matching Recall (\%)}\\
\cmidrule{1-12}
& Benchmark & \multicolumn{5}{c|}{3DMatch} & \multicolumn{5}{c}{3DLoMatch} \\
& Samples & 5000 & 2500 & 1000 & 500 & 250 & 5000 & 2500 & 1000 & 500 & 250 \\
\cmidrule{1-12}
\multirow{6}{*}{\rotatebox[origin=c]{90}{descriptor-based}}
& PerfectMatch\cite{gojcic2019perfect}
& 95.0 & 94.3 & 92.9 & 90.1 & 82.9
& 63.6 & 61.7 & 53.6 & 45.2 & 34.2\\
& FCGF\cite{choy2019fcgf}
& 97.4 & 97.3 & 97.0 & 96.7 & 96.6
& 76.6 & 75.4 & 74.2 & 71.7 & 67.3\\
& D3Feat\cite{bai2020d3feat}
& 95.6 & 95.4 & 94.5 & 94.1 & 93.1
& 67.3 & 66.7 & 67.0 & 66.7 & 66.5\\
& SpinNet\cite{ao2021spinnet}
& 97.6 & 97.2 & 96.8 & 95.5 & 94.3
& 75.3 & 74.9 & 72.5 & 70.0 & 63.6\\
& YOHO\cite{wang2022yoho}
& 98.2 & 97.6 & 97.5 & 97.7 & 96.0
& 79.4 & 78.1 & 76.3 & 73.8 & 69.1\\
& Predator\cite{huang2021predator}
& 96.6 & 96.6 & 96.5 & 96.3 & 96.5
& 78.6 & 77.4 & 76.3 & 75.7 & 75.3\\
\cmidrule{1-12}
\multirow{8}{*}{\rotatebox[origin=c]{90}{correspondence-based}}
& CoFiNet\cite{yu2021cofinet}
& 98.1 & 98.3 & 98.1 & 98.2 & 98.3
& 83.1 & 83.5 & 83.3 & 83.1 & 82.6\\
& GeoTransformer\cite{qin2023geotransformer}
& 97.9 & 97.9 & 97.9 & 97.9 & 97.6
& 88.3 & 88.6 & 88.8 & 88.6 & 88.3\\
& OIF-Net~\cite{yang2022one}
& 98.1 & 98.1 & 97.9 & 98.4 & 98.4
& 84.6 & 85.2 & 85.5 & 86.6 & 87.0\\
& RoITr~\cite{yu2023roitr}
& 98.0 & 98.0 & 97.9 & 98.0 & 97.9
& \textbf{89.6} & \textbf{89.6} & \textbf{89.5} & \textbf{89.4} & \textbf{89.3}\\
& PEAL~\cite{yu2023peal}
& \textbf{98.5} & \textbf{98.6} & \textbf{98.6} & \textbf{98.7} & \textbf{98.7}
& 89.1 & 89.2 & 89.0 & 89.0 & 88.8\\
& SIRA-PCR~\cite{chen2023sira}
& 98.2 & 98.4 & 98.4 & 98.5 & 98.5
& 88.8 & 89.0 & 88.9 & 88.6 & 87.7\\
& DiffusionPCR~\cite{chen2023diffusionpcr}
& 98.3 & 98.3 & 98.3 & 98.3 & 98.3
& 86.3 & 85.9 & 86.0 & 86.1 & 85.9\\
& CAST
& - & 98.3 & 98.3 & 98.4 & 98.3
& - & 83.1 & 83.6 & 85.5 & 84.7 \\
\bottomrule
\end{tabular}
\vspace{-3mm}
\end{table}

\paragraph{Indoor Registration Performance.}
We demonstrate the accuracy of CAST for indoor RGB-D point cloud registration by comparing it with various point cloud registration methods~\cite{fischler1981ransac,yang2021teaser,chen2022sc2,choy2020dgr,bai2021pointdsc,zhang2023mac,zhang2024fastmac} in Table~\ref{table:reg}. All of the registration methods leverage the prevalent FCGF descriptor~\cite{choy2019fcgf}, and FastMAC~\cite{zhang2024fastmac} uses a sampling ratio of 50\%.
For a fair comparison, we follow the evaluation strategy of MAC~\cite{zhang2023mac} to re-compute the registration recall of our method, which is formulated as the fraction of point cloud pairs with RTE$<$30cm and RRE$<$15°. Our method achieves the highest registration recall and the lowest registration errors, suggesting its robustness and accuracy.

\paragraph{Generalization Studies.} To extensively evaluate the generalizability of the proposed CAST in unseen domains, we conduct a generalization experiment from the outdoor dataset KITTI~\cite{geiger2012kitti} to another outdoor dataset ETH~\cite{Pomerleau2012eth}. Note that the KITTI and ETH datasets use Velodyne-64 3D LiDAR and Hokuyo 2D LiDAR, respectively, leading to very different appearances and distributions of point clouds. Hence, our generalization study is practical in applications and solid to demonstrate the generalizability of different methods. For fairness, all methods adopt 30cm for voxel down-sampling, and all methods involving RANSAC set the maximum iterations to be 50000 and the confidence to be 0.999 as the convergence criteria. To enhance the robustness, our method is combined with RANSAC estimating an initial pose from 250 coarse correspondences to reject the outliers during fine matching, and utilizes the global registration in GeoTransformer~\cite{qin2023geotransformer} to refine the pose estimate.

We present the translation errors, rotation errors, and the registration recalls in Table~\ref{table:eth}. Our method achieves satisfying accuracy and robustness, showcasing better generalizability than the coarse-to-fine baseline GeoTransformer~\cite{qin2023geotransformer} and other point-wise descriptors~\cite{choy2019fcgf,huang2024kdd,huang2021predator}. We also compare the our learnable compatibility graph embedding (CGE) with spectral matching (SM)~\cite{2005spectral} for outlier rejection in our method. With RANSAC filtering out severe outliers in advance, spectral matching can lead to better performance than learning-based CGE in unseen domains. Notably, all point-wise methods including CAST exhibit lower registration recalls in generalization studies than patch-wise local descriptor SpinNet~\cite{ao2021spinnet} and BUFFER~\cite{ao2023buffer} incorporating patch-wise and point-wise features. This is mainly because they adopt a feature pyramid network architecture to learn features with abundant global context, which is detrimental for generalization~\cite{ao2021spinnet}. Furthermore, we conduct an unsupervised domain adaptation (UDA) experiment for CAST, which tunes the network by learning to align a point cloud to itself after random rotation and cropping. The results indicate that our model can easily adapt to an unseen domain and achieve robust and accurate performance after one epoch's unsupervised tuning (only 20 minutes on an NVIDIA RTX3090 GPU).

\paragraph{Experiment Statistical Significance.}
Finally, Table~\ref{tab:var} reports the standard deviations (1-sigma) of our evaluation metrics, which are assumed to be Gaussian distributed.
Despite the randomness from voxel down-sampling and RANSAC, the performance of our method remains stable.
Notably, the runtime of some methods such as~\cite{choy2019fcgf,ao2021spinnet,wang2022yoho} in Table~\ref{table:3dmatch} are quite different from results in~\cite{ao2023buffer}, since we report the average runtime including data preprocessing, feature extraction, feature matching, and pose estimation, while the source codes of these methods save some intermediate results such as descriptors to avoid repeated calculation of the same point cloud in different pairs, which leads to unfair runtime comparison.

\begin{table}
\footnotesize
\caption{Registration results on indoor RGBD point cloud datasets.}
\label{table:reg}
\centering
\setlength\tabcolsep{4pt}
\begin{tabular}{c|ccc|ccc}
\toprule
\multirow{2}{*}{Methods} & \multicolumn{3}{c|}{3DMatch} & \multicolumn{3}{c}{3DLoMatch} \\
& RR (\%) & RTE (cm) & RRE (°) & RR (\%) & RTE (cm) & RRE (°) \\
\cmidrule{1-7}
RANSAC-1M~\cite{fischler1981ransac} & 88.42 & 9.42 & 3.05 & 9.77 & 14.87 & 7.01 \\
RANSAC-4M~\cite{fischler1981ransac} & 91.44 & 8.38 & 2.69 &10.44 & 15.14 & 6.91 \\
TEASER++~\cite{yang2021teaser} & 85.77 & 8.66 & 2.73 & 46.76 & 12.89 & 4.12 \\
SC$^2$-PCR~\cite{chen2022sc2} & 93.16 & 6.51 & 2.09 & 58.73 & 10.44 & 3.80 \\
DGR~\cite{choy2020dgr} & 88.85 & 7.02 & 2.28 & 43.80 & 10.82 & 4.17 \\
PointDSC~\cite{bai2021pointdsc} & 91.87 & 6.54 & 2.10 & 56.20 & 10.48 & 3.87 \\
MAC~\cite{zhang2023mac} & 93.72 & 6.54 & 2.02 & 59.85 & 9.75 & 3.50 \\
FastMAC~\cite{zhang2024fastmac} & 92.67 & 6.47 & 2.00 & 58.23 & 10.81 & 3.80\\
CAST & \textbf{96.48} & \textbf{5.64} & \textbf{1.71} & \textbf{76.13} & \textbf{8.47} & \textbf{2.75} \\
\bottomrule
\end{tabular}
\end{table}

\begin{table}
\footnotesize
\centering
\begin{tabular}{c|ccccc|ccc|ccc}
\toprule
Dataset & \multicolumn{5}{c|}{3DMatch} & \multicolumn{3}{c|}{KITTI} & \multicolumn{3}{c}{nuScenes}\\
Metrics & RR & IR & FMR & PIR & PMR & RR & RTE & RRE & RR & RTE & RRE\\
\cmidrule{1-12}
STD & 0.4\% & 0.4\% & 0.5\% & 0.3\% & 0.2\% & 0.0\% & 0.1cm & 0.01° & 0.0\% & 0.1cm & 0.01°\\
\bottomrule
\end{tabular}
\caption{Empirical standard deviations of the evaluation metrics of CAST in repeated experiments.}
\label{tab:var}
\end{table}

\begin{table}
\footnotesize
\caption{Results of generalization from KITTI to ETH.}
\vspace{-2mm}
\label{table:eth}
\centering
\setlength\tabcolsep{4pt}
\begin{tabular}{c|ccc}
\toprule
Methods & RTE (cm) & RRE (°) & RR (\%) \\
\cmidrule{1-4}
FCGF~\cite{choy2019fcgf} & 9.08 & 0.94 & 45.86 \\
Predator~\cite{huang2021predator} & 11.72 & 1.38 & 65.64 \\
SpinNet~\cite{ao2021spinnet} & 6.05 & 0.98 & {\blue 99.44} \\
TCKDD~\cite{huang2024kdd} & 9.61 & 0.88 & 92.43 \\
GeoTransformer~\cite{qin2023geotransformer} & {\red 5.97} & 0.73 & 91.87 \\
BUFFER~\cite{ao2023buffer} & {\blue 6.02} & 0.71 & \textbf{100.00} \\
\cmidrule{1-4}
CAST (CGE) & 6.85 & {\blue 0.65} & 97.76 \\
CAST (SM) & 6.66 & {\red 0.61} & 98.04 \\
CAST + UDA & \textbf{5.25} & \textbf{0.56} & {\red 99.58} \\
\bottomrule
\end{tabular}
\vspace{-2mm}
\end{table}

\subsection{Qualitative Results}
\label{subsec:qualitative}
\begin{figure}
\centering
\includegraphics[width=\linewidth]{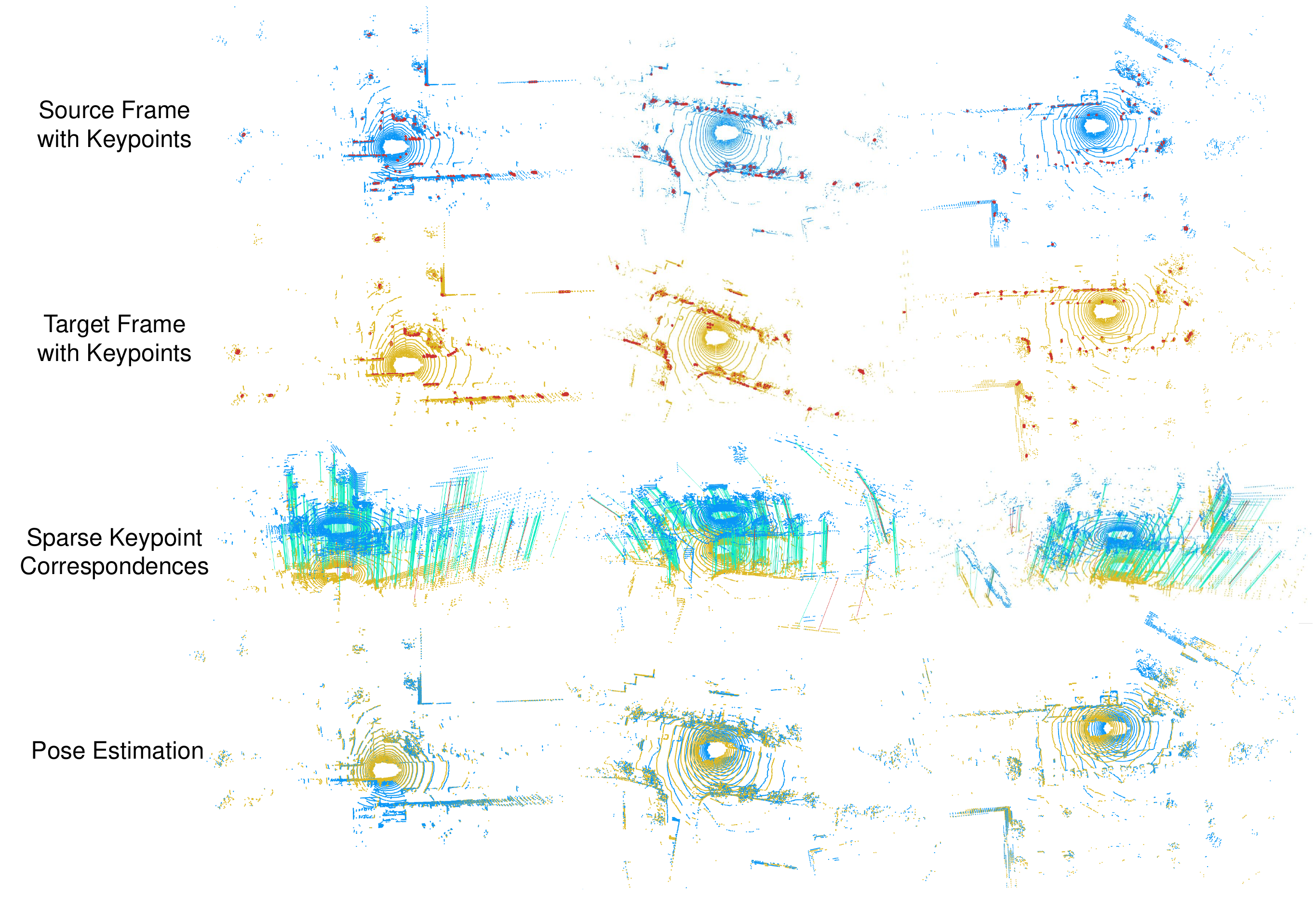}
\caption{Qualitative registration results on nuScenes dataset.
We show three examples in three columns to demonstrate the effectness of CAST in keypoint extraction, matching, and pose estimation.
}
\label{fig:nuscenes}
\end{figure}

\begin{figure}[H]
\centering
\includegraphics[width=\linewidth]{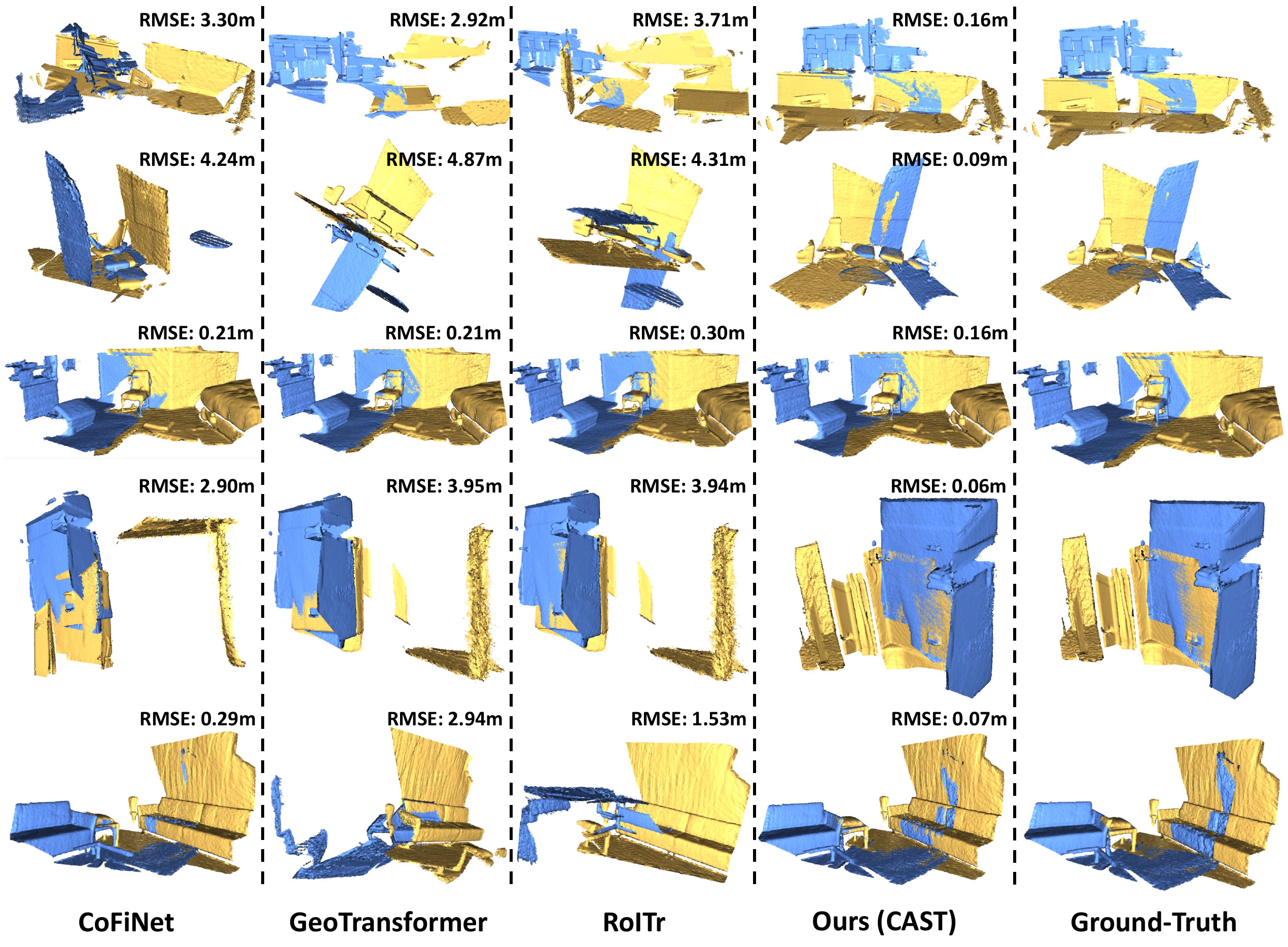}
\caption{Qualitative registration results of CoFiNet~\cite{yu2021cofinet}, GeoTransformer~\cite{qin2023geotransformer}, RoITr~\cite{yu2023roitr}, and CAST compared with the ground truth alignment on 3DMatch dataset. We present five examples in five rows, which demonstrate the robustness and accuracy of our method.}
\label{fig:3dmatch}
\end{figure}

Figure~\ref{fig:kitti}, Figure~\ref{fig:nuscenes}, and Figure~\ref{fig:3dmatch} provide qualitative results about the registration performance on outdoor datasets KITTI~\cite{geiger2012kitti}, nuScenes~\cite{nuscenes2019}, and the indoor dataset 3DMatch~\cite{zeng20173dmatch}, respectively.

\subsection{Limitation}
\label{subsec:limitation}
The main limitation of the proposed CAST is the sub-optimal performance in low overlapping scenarios such as the 3DLoMatch benchmark compared to \textit{state-of-the-art} methods, which may be ascribed to two aspects.
(1) There is no effective outlier rejection for coarse matching as it is difficult to search inliers from patch correspondences based on geometry consistency due to low resolution.
(2) Due to the sparsity and non-uniformity of keypoints, the inlier ratio of keypoint correspondences still falls short of what is required for robust pose estimation without a hypothesis-and-selection pipeline.
Nevertheless, this RANSAC-free lightweight fine matching pipeline can achieve satisfying performance in outdoor scenarios.
Considering the superior PIR and PMR of our coarse matching, we may directly exploit dense feature matching to enhance the robustness in low overlapping point cloud registration scenarios as a future work.

\subsection{Broader Impacts}
\label{subsec:impact}
We present a novel consistency-aware spot-guided Transformer based on sparse attention to extract consistent coarse correspondences from point clouds.
In addition, we propose a lightweight fine matching module for versatile and hierarchical point cloud registration, benefiting from the efficiency of sparse keypoint matching and the accuracy of dense registration.
Different from existing methods, our fine matching is based on flexible local attention instead of optimal transport heavily relying on patch-to-patch correspondences, thus allowing independent deployment without coarse matching.
Besides, the sparsity of keypoints ensures the efficiency of spatial consistency filtering.

Enjoying these merits, this work not only achieves superior accuracy, efficiency, and robustness in point cloud registration, but also paves the way to various large-scale real-time applications, such as SfM, SLAM, autonomous driving, or any other where point cloud registration plays a role.
For examples, the reconstruction of indoor scenes and objects from unlabeled 3D scans could benefit from our work, which can precisely recover the rigid transformation between different scans.
Additionally, our fine matching may independently construct a real-time LiDAR-based or RGBD camera-based odometry system for SLAM or SfM, as it is capable of efficient and reliable local data association and accurate pose estimation between two large-scale point clouds with a strong pose prior, while our coarse matching could be utilized in place recognition and global re-localization in SLAM.

As our work aims at tackling a fundamental problem in 3D computer vision, we do not anticipate a direct negative impact. Potential negative outcomes might occur in real applications where our method is involved.

\newpage

\newpage
\section*{NeurIPS Paper Checklist}

\begin{enumerate}

\item {\bf Claims}
    \item[] Question: Do the main claims made in the abstract and introduction accurately reflect the paper's contributions and scope?
    \item[] Answer: \answerYes{}
    \item[] Justification: Our abstract and introduction clearly state the claims made, including the motivations, contributions, and the performance of our approach.
    \item[] Guidelines:
    \begin{itemize}
        \item The answer NA means that the abstract and introduction do not include the claims made in the paper.
        \item The abstract and/or introduction should clearly state the claims made, including the contributions made in the paper and important assumptions and limitations. A No or NA answer to this question will not be perceived well by the reviewers. 
        \item The claims made should match theoretical and experimental results, and reflect how much the results can be expected to generalize to other settings. 
        \item It is fine to include aspirational goals as motivation as long as it is clear that these goals are not attained by the paper. 
    \end{itemize}

\item {\bf Limitations}
    \item[] Question: Does the paper discuss the limitations of the work performed by the authors?
    \item[] Answer: \answerYes{}
    \item[] Justification: We have detailed the limitations of our work in Sec.~\ref{subsec:limitation} of the appendix, and we also point out the scope of our claims made, including the benchmarks and sensors (Sec.~\ref{subsec:data}) and the computational efficiency (Sec.~\ref{sec:experiment}), \textit{etc}.
    \item[] Guidelines:
    \begin{itemize}
        \item The answer NA means that the paper has no limitation while the answer No means that the paper has limitations, but those are not discussed in the paper. 
        \item The authors are encouraged to create a separate "Limitations" section in their paper.
        \item The paper should point out any strong assumptions and how robust the results are to violations of these assumptions (e.g., independence assumptions, noiseless settings, model well-specification, asymptotic approximations only holding locally). The authors should reflect on how these assumptions might be violated in practice and what the implications would be.
        \item The authors should reflect on the scope of the claims made, e.g., if the approach was only tested on a few datasets or with a few runs. In general, empirical results often depend on implicit assumptions, which should be articulated.
        \item The authors should reflect on the factors that influence the performance of the approach. For example, a facial recognition algorithm may perform poorly when image resolution is low or images are taken in low lighting. Or a speech-to-text system might not be used reliably to provide closed captions for online lectures because it fails to handle technical jargon.
        \item The authors should discuss the computational efficiency of the proposed algorithms and how they scale with dataset size.
        \item If applicable, the authors should discuss possible limitations of their approach to address problems of privacy and fairness.
        \item While the authors might fear that complete honesty about limitations might be used by reviewers as grounds for rejection, a worse outcome might be that reviewers discover limitations that aren't acknowledged in the paper. The authors should use their best judgment and recognize that individual actions in favor of transparency play an important role in developing norms that preserve the integrity of the community. Reviewers will be specifically instructed to not penalize honesty concerning limitations.
    \end{itemize}

\item {\bf Theory Assumptions and Proofs}
    \item[] Question: For each theoretical result, does the paper provide the full set of assumptions and a complete (and correct) proof?
    \item[] Answer: \answerNA{}
    \item[] Justification: This paper does not include any theoretical results.
    \item[] Guidelines:
    \begin{itemize}
        \item The answer NA means that the paper does not include theoretical results. 
        \item All the theorems, formulas, and proofs in the paper should be numbered and cross-referenced.
        \item All assumptions should be clearly stated or referenced in the statement of any theorems.
        \item The proofs can either appear in the main paper or the supplemental material, but if they appear in the supplemental material, the authors are encouraged to provide a short proof sketch to provide intuition. 
        \item Inversely, any informal proof provided in the core of the paper should be complemented by formal proofs provided in appendix or supplemental material.
        \item Theorems and Lemmas that the proof relies upon should be properly referenced. 
    \end{itemize}

    \item {\bf Experimental Result Reproducibility}
    \item[] Question: Does the paper fully disclose all the information needed to reproduce the main experimental results of the paper to the extent that it affects the main claims and/or conclusions of the paper (regardless of whether the code and data are provided or not)?
    \item[] Answer: \answerYes{}
    \item[] Justification: This paper fully discloses all the information needed to reproduce the main experimental results, including the detailed system architecture in Sec.~\ref{sec:method} and some modular architectures along with the hyper-parameters in Sec.~\ref{subsec:architecture}, the training and evaluation settings in Sec.~\ref{sec:experiment}, and datasets with metrics in Sec.~\ref{subsec:data} and Sec.~\ref{subsec:metric}, respectively.
    \item[] Guidelines:
    \begin{itemize}
        \item The answer NA means that the paper does not include experiments.
        \item If the paper includes experiments, a No answer to this question will not be perceived well by the reviewers: Making the paper reproducible is important, regardless of whether the code and data are provided or not.
        \item If the contribution is a dataset and/or model, the authors should describe the steps taken to make their results reproducible or verifiable. 
        \item Depending on the contribution, reproducibility can be accomplished in various ways. For example, if the contribution is a novel architecture, describing the architecture fully might suffice, or if the contribution is a specific model and empirical evaluation, it may be necessary to either make it possible for others to replicate the model with the same dataset, or provide access to the model. In general. releasing code and data is often one good way to accomplish this, but reproducibility can also be provided via detailed instructions for how to replicate the results, access to a hosted model (e.g., in the case of a large language model), releasing of a model checkpoint, or other means that are appropriate to the research performed.
        \item While NeurIPS does not require releasing code, the conference does require all submissions to provide some reasonable avenue for reproducibility, which may depend on the nature of the contribution. For example
        \begin{enumerate}
            \item If the contribution is primarily a new algorithm, the paper should make it clear how to reproduce that algorithm.
            \item If the contribution is primarily a new model architecture, the paper should describe the architecture clearly and fully.
            \item If the contribution is a new model (e.g., a large language model), then there should either be a way to access this model for reproducing the results or a way to reproduce the model (e.g., with an open-source dataset or instructions for how to construct the dataset).
            \item We recognize that reproducibility may be tricky in some cases, in which case authors are welcome to describe the particular way they provide for reproducibility. In the case of closed-source models, it may be that access to the model is limited in some way (e.g., to registered users), but it should be possible for other researchers to have some path to reproducing or verifying the results.
        \end{enumerate}
    \end{itemize}

\item {\bf Open access to data and code}
    \item[] Question: Does the paper provide open access to the data and code, with sufficient instructions to faithfully reproduce the main experimental results, as described in supplemental material?
    \item[] Answer: \answerYes{}
    \item[] Justification: We have submitted the data and codes in the supplementary material with detailed instructions on data access and preparation as well as guidelines to reproduce all experimental results. The paper will provide public access to the codes upon acceptance.
    \item[] Guidelines:
    \begin{itemize}
        \item The answer NA means that paper does not include experiments requiring code.
        \item Please see the NeurIPS code and data submission guidelines (\url{https://nips.cc/public/guides/CodeSubmissionPolicy}) for more details.
        \item While we encourage the release of code and data, we understand that this might not be possible, so “No” is an acceptable answer. Papers cannot be rejected simply for not including code, unless this is central to the contribution (e.g., for a new open-source benchmark).
        \item The instructions should contain the exact command and environment needed to run to reproduce the results. See the NeurIPS code and data submission guidelines (\url{https://nips.cc/public/guides/CodeSubmissionPolicy}) for more details.
        \item The authors should provide instructions on data access and preparation, including how to access the raw data, preprocessed data, intermediate data, and generated data, etc.
        \item The authors should provide scripts to reproduce all experimental results for the new proposed method and baselines. If only a subset of experiments are reproducible, they should state which ones are omitted from the script and why.
        \item At submission time, to preserve anonymity, the authors should release anonymized versions (if applicable).
        \item Providing as much information as possible in supplemental material (appended to the paper) is recommended, but including URLs to data and code is permitted.
    \end{itemize}

\item {\bf Experimental Setting/Details}
    \item[] Question: Does the paper specify all the training and test details (e.g., data splits, hyperparameters, how they were chosen, type of optimizer, etc.) necessary to understand the results?
    \item[] Answer: \answerYes{}
    \item[] Justification: This paper fully specifies all the training and evaluation settings, including the optimizer and the learning rate scheduler in Sec.~\ref{sec:experiment}, hyper-parameters in Sec.~\ref{sec:appendix}, data splits and preparation in Sec.~\ref{subsec:data}.
    \item[] Guidelines:
    \begin{itemize}
        \item The answer NA means that the paper does not include experiments.
        \item The experimental setting should be presented in the core of the paper to a level of detail that is necessary to appreciate the results and make sense of them.
        \item The full details can be provided either with the code, in appendix, or as supplemental material.
    \end{itemize}

\item {\bf Experiment Statistical Significance}
    \item[] Question: Does the paper report error bars suitably and correctly defined or other appropriate information about the statistical significance of the experiments?
    \item[] Answer: \answerYes{}
    \item[] Justification: In this paper, all metrics are assumed to be normally distributed, whose empirical standard deviations (1-sigma) in repeated experiments are reported in Table~\ref{tab:var} to demonstrate the experiment statistical significance. The randomness mainly comes from voxel down-sampling and RANSAC.
    \item[] Guidelines:
    \begin{itemize}
        \item The answer NA means that the paper does not include experiments.
        \item The authors should answer "Yes" if the results are accompanied by error bars, confidence intervals, or statistical significance tests, at least for the experiments that support the main claims of the paper.
        \item The factors of variability that the error bars are capturing should be clearly stated (for example, train/test split, initialization, random drawing of some parameter, or overall run with given experimental conditions).
        \item The method for calculating the error bars should be explained (closed form formula, call to a library function, bootstrap, etc.)
        \item The assumptions made should be given (e.g., Normally distributed errors).
        \item It should be clear whether the error bar is the standard deviation or the standard error of the mean.
        \item It is OK to report 1-sigma error bars, but one should state it. The authors should preferably report a 2-sigma error bar than state that they have a 96\% CI, if the hypothesis of Normality of errors is not verified.
        \item For asymmetric distributions, the authors should be careful not to show in tables or figures symmetric error bars that would yield results that are out of range (e.g. negative error rates).
        \item If error bars are reported in tables or plots, The authors should explain in the text how they were calculated and reference the corresponding figures or tables in the text.
    \end{itemize}

\item {\bf Experiments Compute Resources}
    \item[] Question: For each experiment, does the paper provide sufficient information on the computer resources (type of compute workers, memory, time of execution) needed to reproduce the experiments?
    \item[] Answer: \answerYes{}
    \item[] Justification: In Sec.~\ref{sec:experiment}, we have detailed the type of CPU and GPU of our device, and the runtime of the proposed method and nearly all baselines on our computer.
    \item[] Guidelines:
    \begin{itemize}
        \item The answer NA means that the paper does not include experiments.
        \item The paper should indicate the type of compute workers CPU or GPU, internal cluster, or cloud provider, including relevant memory and storage.
        \item The paper should provide the amount of compute required for each of the individual experimental runs as well as estimate the total compute. 
        \item The paper should disclose whether the full research project required more compute than the experiments reported in the paper (e.g., preliminary or failed experiments that didn't make it into the paper). 
    \end{itemize}
    
\item {\bf Code Of Ethics}
    \item[] Question: Does the research conducted in the paper conform, in every respect, with the NeurIPS Code of Ethics \url{https://neurips.cc/public/EthicsGuidelines}?
    \item[] Answer: \answerYes{}
    \item[] Justification: This research conforms with the NeurIPS Code of Ethics in every respect.
    \item[] Guidelines:
    \begin{itemize}
        \item The answer NA means that the authors have not reviewed the NeurIPS Code of Ethics.
        \item If the authors answer No, they should explain the special circumstances that require a deviation from the Code of Ethics.
        \item The authors should make sure to preserve anonymity (e.g., if there is a special consideration due to laws or regulations in their jurisdiction).
    \end{itemize}

\item {\bf Broader Impacts}
    \item[] Question: Does the paper discuss both potential positive societal impacts and negative societal impacts of the work performed?
    \item[] Answer: \answerYes{}
    \item[] Justification: The broader impacts are discussed in Sec.~\ref{subsec:impact} of the appendix.
    \item[] Guidelines:
    \begin{itemize}
        \item The answer NA means that there is no societal impact of the work performed.
        \item If the authors answer NA or No, they should explain why their work has no societal impact or why the paper does not address societal impact.
        \item Examples of negative societal impacts include potential malicious or unintended uses (e.g., disinformation, generating fake profiles, surveillance), fairness considerations (e.g., deployment of technologies that could make decisions that unfairly impact specific groups), privacy considerations, and security considerations.
        \item The conference expects that many papers will be foundational research and not tied to particular applications, let alone deployments. However, if there is a direct path to any negative applications, the authors should point it out. For example, it is legitimate to point out that an improvement in the quality of generative models could be used to generate deepfakes for disinformation. On the other hand, it is not needed to point out that a generic algorithm for optimizing neural networks could enable people to train models that generate Deepfakes faster.
        \item The authors should consider possible harms that could arise when the technology is being used as intended and functioning correctly, harms that could arise when the technology is being used as intended but gives incorrect results, and harms following from (intentional or unintentional) misuse of the technology.
        \item If there are negative societal impacts, the authors could also discuss possible mitigation strategies (e.g., gated release of models, providing defenses in addition to attacks, mechanisms for monitoring misuse, mechanisms to monitor how a system learns from feedback over time, improving the efficiency and accessibility of ML).
    \end{itemize}
    
\item {\bf Safeguards}
    \item[] Question: Does the paper describe safeguards that have been put in place for responsible release of data or models that have a high risk for misuse (e.g., pretrained language models, image generators, or scraped datasets)?
    \item[] Answer: \answerNA{}
    \item[] Justification: This paper poses no such risks for misuse.
    \item[] Guidelines:
    \begin{itemize}
        \item The answer NA means that the paper poses no such risks.
        \item Released models that have a high risk for misuse or dual-use should be released with necessary safeguards to allow for controlled use of the model, for example by requiring that users adhere to usage guidelines or restrictions to access the model or implementing safety filters. 
        \item Datasets that have been scraped from the Internet could pose safety risks. The authors should describe how they avoided releasing unsafe images.
        \item We recognize that providing effective safeguards is challenging, and many papers do not require this, but we encourage authors to take this into account and make a best faith effort.
    \end{itemize}

\item {\bf Licenses for existing assets}
    \item[] Question: Are the creators or original owners of assets (e.g., code, data, models), used in the paper, properly credited and are the license and terms of use explicitly mentioned and properly respected?
    \item[] Answer: \answerYes{}
    \item[] Justification: This paper involves existing benchmarks and utilizes existing methods for evaluation, whose papers are properly cited with licenses. The related URL are included in our codes.
    \item[] Guidelines:
    \begin{itemize}
        \item The answer NA means that the paper does not use existing assets.
        \item The authors should cite the original paper that produced the code package or dataset.
        \item The authors should state which version of the asset is used and, if possible, include a URL.
        \item The name of the license (e.g., CC-BY 4.0) should be included for each asset.
        \item For scraped data from a particular source (e.g., website), the copyright and terms of service of that source should be provided.
        \item If assets are released, the license, copyright information, and terms of use in the package should be provided. For popular datasets, \url{paperswithcode.com/datasets} has curated licenses for some datasets. Their licensing guide can help determine the license of a dataset.
        \item For existing datasets that are re-packaged, both the original license and the license of the derived asset (if it has changed) should be provided.
        \item If this information is not available online, the authors are encouraged to reach out to the asset's creators.
    \end{itemize}

\item {\bf New Assets}
    \item[] Question: Are new assets introduced in the paper well documented and is the documentation provided alongside the assets?
    \item[] Answer: \answerYes{}
    \item[] Justification: This paper provides details of our model in both main part (Sec.~\ref{sec:method}) and the appendix (Sec.~\ref{subsec:architecture}). Our source codes have been submitted in the supplementary material while no new datasets are proposed.
    \item[] Guidelines:
    \begin{itemize}
        \item The answer NA means that the paper does not release new assets.
        \item Researchers should communicate the details of the dataset/code/model as part of their submissions via structured templates. This includes details about training, license, limitations, etc. 
        \item The paper should discuss whether and how consent was obtained from people whose asset is used.
        \item At submission time, remember to anonymize your assets (if applicable). You can either create an anonymized URL or include an anonymized zip file.
    \end{itemize}

\item {\bf Crowdsourcing and Research with Human Subjects}
    \item[] Question: For crowdsourcing experiments and research with human subjects, does the paper include the full text of instructions given to participants and screenshots, if applicable, as well as details about compensation (if any)? 
    \item[] Answer: \answerNA{}
    \item[] Justification: This paper does not involve crowdsourcing nor research with human subjects.
    \item[] Guidelines:
    \begin{itemize}
        \item The answer NA means that the paper does not involve crowdsourcing nor research with human subjects.
        \item Including this information in the supplemental material is fine, but if the main contribution of the paper involves human subjects, then as much detail as possible should be included in the main paper. 
        \item According to the NeurIPS Code of Ethics, workers involved in data collection, curation, or other labor should be paid at least the minimum wage in the country of the data collector. 
    \end{itemize}

\item {\bf Institutional Review Board (IRB) Approvals or Equivalent for Research with Human Subjects}
    \item[] Question: Does the paper describe potential risks incurred by study participants, whether such risks were disclosed to the subjects, and whether Institutional Review Board (IRB) approvals (or an equivalent approval/review based on the requirements of your country or institution) were obtained?
    \item[] Answer: \answerNA{}
    \item[] Justification: This paper does not involve crowdsourcing nor research with human subjects.
    \item[] Guidelines:
    \begin{itemize}
        \item The answer NA means that the paper does not involve crowdsourcing nor research with human subjects.
        \item Depending on the country in which research is conducted, IRB approval (or equivalent) may be required for any human subjects research. If you obtained IRB approval, you should clearly state this in the paper. 
        \item We recognize that the procedures for this may vary significantly between institutions and locations, and we expect authors to adhere to the NeurIPS Code of Ethics and the guidelines for their institution. 
        \item For initial submissions, do not include any information that would break anonymity (if applicable), such as the institution conducting the review.
    \end{itemize}

\end{enumerate}

\end{document}